%% file: t1d-main.tex
\documentclass[pmlr]{jmlr}

\usepackage{longtable}

\usepackage{booktabs}
\usepackage{multirow}
\usepackage[load-configurations=version-1]{siunitx} 

\makeatletter
\def\set@curr@file#1{\def\@curr@file{#1}} 
\makeatother

\usepackage{dsfont}
\usepackage{todonotes}
\usepackage{subcaption}

\usepackage{wrapfig}

\usepackage{xspace}
\newcommand{\dtdsim}{\texttt{DTD-Sim}\xspace}
\newcommand{\vv}[1]{\mathbf{#1}}

\theorembodyfont{\upshape}
\theoremheaderfont{\scshape}
\theorempostheader{:}
\theoremsep{\newline}

\input{mathdefs.tex}

\jmlrvolume{126}
\jmlryear{2020}
\jmlrworkshop{Machine Learning for Healthcare}

\title[Learning Insulin-Glucose Dynamics in the Wild]{Learning Insulin-Glucose Dynamics in the Wild}

\author{}
\author{\Name{Andrew C.~Miller} \Email{acmiller@apple.com}
      \addr \\
      Apple \\
      Seattle, WA, USA
      \AND
      \Name{Nicholas J.~Foti} \Email{nicholas\_foti@apple.com}
      \addr \\
      Apple \\
      Seattle, WA, USA
      \AND
      \Name{Emily Fox} \Email{emily\_fox@apple.com}
      \addr \\
      Apple \\
      Seattle, WA, USA
}

\begin{document}
\maketitle
\begin{abstract}
\input{sections/abstract.tex}
\end{abstract}

\input{sections/intro.tex}

\input{sections/cohort.tex}

\input{sections/methods.tex}

\input{sections/experiments.tex}

\input{sections/discussion.tex}


\acks{ACM thanks Ben Kreis, Nate Racklyeft, Jen Block, and Luke Winstrom for supporting this research project, and Leon Gatys for enriching methodological discussions and review of an early draft.  The authors also thank Sean Jewell for review of a later draft. }

\bibliography{refs.bib}

\newpage
\appendix
\input{sections/uva-simulator.tex}
\input{sections/app-experiments.tex}
\input{sections/app-inference.tex}

\end{document}

%% file: mathdefs.tex
\usepackage{tcolorbox}

\newcommand{\given}{\,|\,}

\newcommand{\ba}{\boldsymbol{a}}

\newcommand{\bd}{\boldsymbol{d}}

\newcommand{\bm}{\boldsymbol{m}}

\newcommand{\bp}{\boldsymbol{p}}

\newcommand{\bs}{\boldsymbol{s}}

\newcommand{\bu}{\boldsymbol{u}}

\newcommand{\bx}{\boldsymbol{x}}
\newcommand{\by}{\boldsymbol{y}}
\newcommand{\bz}{\boldsymbol{z}}

\newcommand{\beps}{\boldsymbol{\epsilon}}
\newcommand{\btheta}{\boldsymbol{\theta}}
\newcommand{\bmu}{\boldsymbol{\mu}}

\newcommand{\blambda}{\boldsymbol{\lambda}}
\newcommand{\bSigma}{\boldsymbol{\Sigma}}

%% file: sections/abstract.tex
We develop a new model of insulin-glucose dynamics for forecasting blood glucose in type 1 diabetics.
We augment an existing biomedical model by introducing time-varying dynamics driven by a machine learning sequence model.
Our model maintains a physiologically plausible inductive bias and clinically interpretable parameters --- e.g.,~insulin sensitivity --- while inheriting the flexibility of modern pattern recognition algorithms.
Critical to modeling success are the flexible, but structured representations of subject variability with a sequence model.
In contrast, less constrained models like the LSTM fail to provide reliable or physiologically plausible forecasts.
We conduct an extensive empirical study.  We show that allowing biomedical model dynamics to vary in time improves forecasting at long time horizons, up to six hours, and produces forecasts consistent with the physiological effects of insulin and carbohydrates.

%% file: sections/intro.tex
\section{Introduction}

Type one diabetes (T1D) is an incurable chronic condition in which the pancreas
produces little to no insulin.  This lack of insulin frustrates the regulation
of blood glucose levels.  Left unmanaged, glucose will elevate, ushering in a
host of long- and short-term health complications.
There is no method of prevention, reversal, nor cure; T1D requires constant
management.  Of the estimated 1.25 million Americans with T1D, 75\% are
diagnosed in childhood, resulting in a life-long burden of disease management.
Management typically entails the injection of subcutaneous insulin to regulate
glucose.

T1D is rife with complications.  Insufficient insulin leads
to chronically elevated blood glucose; common complications include
kidney disease, cardiovascular disease, eye disease, and
nerve disease.  Patients diagnosed with T1D before age ten have a thirty-fold
increased risk of coronary heart disease and acute myocardial infarction
compared to matched controls. Early-diagnosed T1D patients face a 14-18 year
loss in life expectancy \citep{rawshani2018excess}.  
Children and
adolescents with T1D begin to show signs of cardiovascular disease after only
ten years of disease duration \citep{singh2003vascular,
jarvisalo2004endothelial, margeirsdottir2010early, rawshani2018excess}.

Excess insulin, on the other hand, can lead to acute complications, such as hypoglycemia.
Too much insulin lowers blood glucose to dangerous levels resulting in loss of
consciousness or even death \citep{snell2012hypoglycemia}.  Existing insulin
delivery systems may target higher than desirable levels of blood glucose to
avoid hypoglycemic events.  This reflects the asymmetry of negative effects of
glucose levels --- chronically elevated glucose has negative long term
consequences, while low glucose levels can be immediately catastrophic.

Such complications can be avoided by delivering ``just enough'' insulin.  
The determination of ``just enough'' at any given moment is a challenge.
One impediment is the unknown (and time-varying) state of the T1D subject ---
e.g.,~How sensitive is she to insulin?  How many grams of carbohydrates has she
absorbed? How do absorbed carbohydrates translate into increased blood glucose?

Patients with T1D endure the constant burden of tuning insulin delivery.
Fewer than one-third of T1D patients in the US consistently achieve target
blood-glucose levels \citep{miller2015current}.  To ease the burden and improve
glucose regulation, automatic insulin delivery systems are becoming the new standard for
T1D management.  Continuous glucose monitors (CGMs) and insulin pumps
facilitate the management of T1D.  Additionally, these devices present the
opportunity to develop more effective insulin delivery algorithms.\footnote{
  Automated delivery systems have been recently developed, measuring glucose every five minutes and automatically adjusting insulin delivery; bolus doses are
  manually requested. See the Tidepool Loop project (\url{https://www.tidepool.org}) for an additional example of an open automated insulin delivery system.
  }

Like manual insulin delivery, automatic systems use CGM and insulin pump
information to determine the appropriate dose of insulin at any given moment.
The insulin pump controller uses forecasts of blood glucose a few hours into the future to deliver the appropriate basal insulin dose. Improved
forecasts enable finer control over glucose levels.

\vspace{-.6em}
\paragraph{Improving forecasts} 
Models of insulin-glucose interaction are used to predict future glucose values.
Traditionally, such models of insulin-glucose dynamics have been based in physiology.
The seminal Bergman \emph{minimal model} is a
multi-compartment insulin model that describes the interaction between
active insulin, blood glucose, and endogenous insulin production
 \citep{bergman1981physiologic, bergman2005minimal}.
The more sophisticated UVA/Padova T1D simulator includes a model for oral ingestion of carbohydrates to describe changes in 
blood glucose \citep{man2014uva}.
The UVA/Padova simulator is a useful tool for validating empirical models ---
it is FDA approved to test the efficacy and risk of insulin
delivery policies in automated delivery systems.
These simulators, however, were developed to be accurate in highly
controlled experimental settings, not for modeling real-world data with noise and missingness. 

In contrast, data driven approaches --- both traditional statistical methods and machine learning tools --- offer the promise of uncovering and leveraging patterns found in the data.  The challenge, of course, is to find models that can infer the intricate (and unobserved) dynamics of the observed data. 

\vspace{-.6em}
\paragraph{Contributions and generalizable insights}
In this work, we develop a hybrid statistical and physiological model 
of insulin-glucose dynamics for producing long-term forecasts from real-world
T1D management data --- CGM, insulin pump, and carbohydrate logs.
Our model strikes a balance between purely statistical and purely physiological
approaches. We show that statistical machine learning model components --- e.g.,~neural networks and state-space models --- can be part of a larger, physiologically-grounded model. This fused model inherits the realistic inductive biases from the physiological model and the flexibility and predictive power of modern machine learning sequence methods.

We show that this hybrid approach can improve forecasts over purely mechanistic or purely statistical approaches on real-world T1D data.  Additionally, we show that our model produces physiologically plausible counterfactual predictions under alternative insulin and meal schedules, whereas statistical approaches do not.
Importantly, we do not claim that our model \emph{solves} glucose forecasting for T1D management --- all models struggle with forecast accuracy at long time horizons.
Rather, our contribution is the first in a new family of hybrid statistical-and-physiological models and evidence that this approach can better describe long term structure in real-world T1D data, an important step toward better management of T1D.

The biomedical and epidemiological literature is replete with structured models of complex biological phenomena that may be too inflexible or under-specified to use with real-world sensor data, but nevertheless provide a useful inductive bias \citep{anderson1992infectious,dalla2002oral,mcsharry2003dynamical,kotani2005model,trayanova2011whole}.
Our statistical-and-physiological hybrid approach has the potential to generalize to other biomedical applications. 
As such, the presented methodology holds promise in many domains beyond T1D. 

Section~\ref{sec:cohort} details the data and subjects incorporated into this work. Section~\ref{sec:methods} details the proposed hybrid model, including the underlying physiological T1D simulator and statistical time series model. Section~\ref{sec:experiments} describes our experimental setup, evaluation metrics, and empirical results. We conclude with a discussion of related work and future research directions in Section~\ref{sec:discussion}.

%% file: sections/cohort.tex
\section{Cohort and Data}
\label{sec:cohort}

\begin{figure*}[t]
\centering
\includegraphics[width=.9\textwidth]{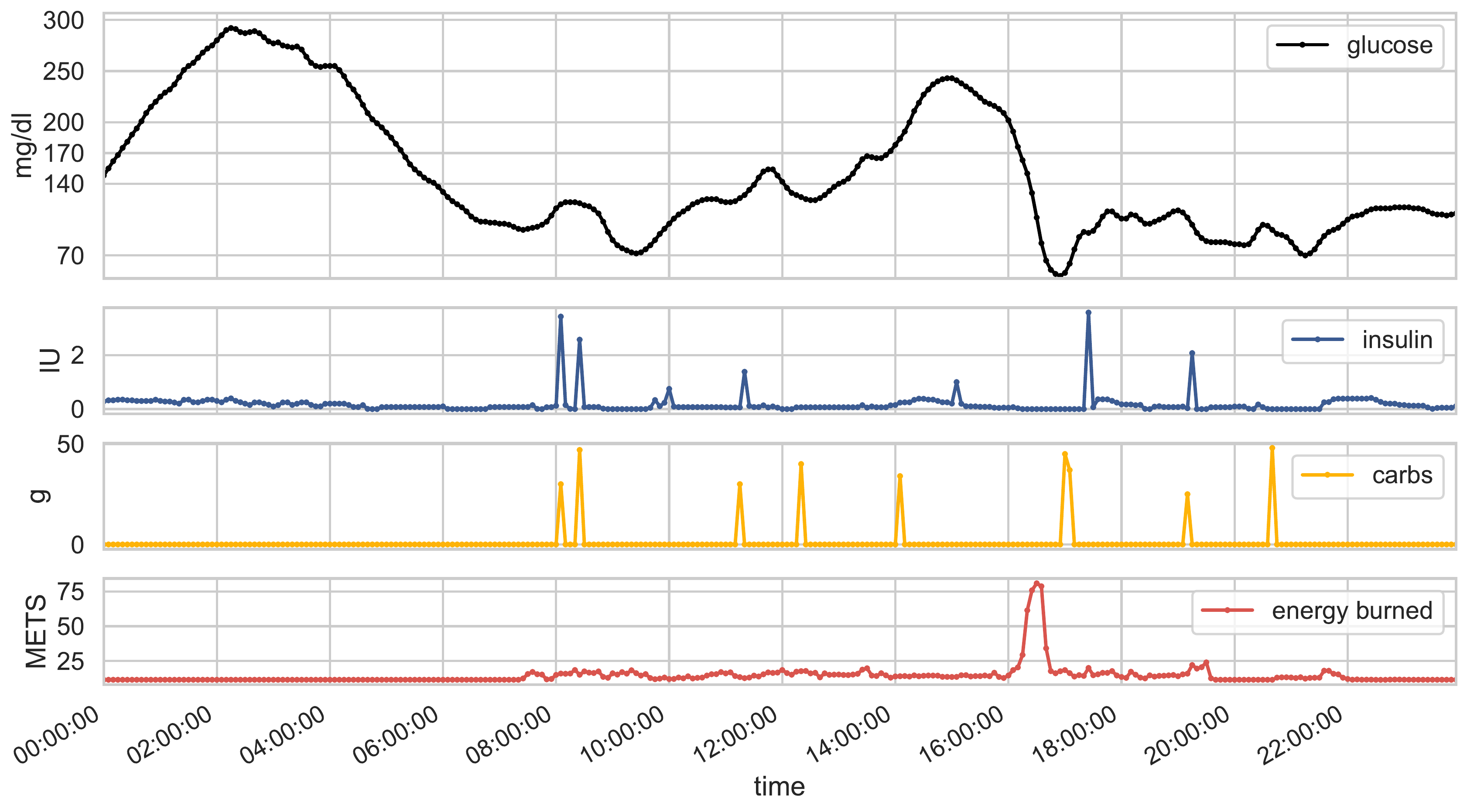}
\caption{Data for a single day: blood glucose as measured by CGM, insulin delivered by pump,
user carbohydrate log, and energy use captured every five minutes by HealthKit. }
\label{fig:example-data}
\end{figure*}

Our data are observational measurements from two T1D participants using a continuous glucose monitor (CGM) and an insulin pump throughout daily life.
The blood glucose level measurements are synchronized and collected using Apple's HealthKit framework.\footnote{\url{https://developer.apple.com/documentation/healthkit}} 
Additionally, we consider the energy spent by the participant throughout the day, as summarized by HealthKit.
For every five minute period the following quantities are recorded:
\vspace{-.5em}
\begin{itemize} \itemsep 0pt
\item instantaneous continuous glucose monitor measurement (mg/dL),
\item basal and bolus insulin delivered (insulin units),
\item estimated carbohydrate ingestion (grams), and
\item estimated active energy burned (METs).
\end{itemize}
\vspace{-.5em}
Figure~\ref{fig:example-data} shows this data for a single day.

The management of T1D is highly personalized --- carbohydrate intake and insulin dosage for one person may not be appropriate for another.  While this work only studies the data of two individuals (separately), we analyze an individual's data streams over a long time window.  We consider data collected over a 150 day period, using the first 120 days to train and validate a forecasting model, and the subsequent 30 days to measure prediction accuracy.  At one sample every five minutes, each participant has recorded over 40{,}000 data points. A summary of the time period for each participant is in Table~\ref{tab:cohort-summary}.

Collecting and processing sensitive health data demands high security standards and strict protocols to protect the privacy and interests of all study participants. The types of data collected --- CGM, insulin pump, meal logs, and activity --- pose a potential privacy concern to participants. 
Upon enrollment, all participants were made aware of these risks via an informed consent form.
To reduce privacy risks, we restrict the focus of our analysis to a small number of temporal data streams relevant to the management of T1D.  Personally identifying information was separated from study data; participants were assigned unique keys.
Additionally, researchers accessed the data through a secure virtual private network (VPN).
As this study is purely retrospective, there were no direct medical risks to participants throughout the course of the study. 

\begin{table*}[t]
  \centering
  \scalebox{.86}{
    \input{figs/data-figs/cohort-table.tex}
  }
  \caption{Cohort and data summary.}
  \label{tab:cohort-summary}
\end{table*}

\subsection{Data Extraction}
The raw CGM and pump data falls on a potentially irregular grid.  For simplicity, we
interpolate CGM values to a fixed five minute grid, e.g.\ 6:00AM, 6:05AM, etc. 
For each five minute period, we compute total insulin units delivered, grams of carbohydrates ingested, and units of energy burned. 
For CGM gaps longer than five minutes in our observations, we linearly interpolate the values for the relevant five minute periods;\footnote{A cubic spline interpolation was initially used, but discarded because it created glucose values outside of the observed range (and in some instances, negative values).} this method accounted for fewer than 2\% of all observations.

%% file: figs/data-figs/cohort-table.tex
\begin{tabular}{lrrrrrr}
\toprule
\multirow{2}{*}{Subject} &
  \multicolumn{3}{c}{Train} &
  \multicolumn{3}{c}{Test} \\
& {start date} & {\# days} & {\# samples} & {start date} & {\# days} & {\# samples} \\
\midrule
1 & March 24 & 120 & 34381 & July 22 & 31 & 8819 \\ 
2 & February 25 & 120 & 34396 & June 25 & 31 & 8804 \\ 
\end{tabular}

%% file: sections/methods.tex
\section{Methods}
\label{sec:methods}
Insulin and glucose obey complicated unobserved dynamics.  
In T1D, subcutaneous insulin takes time to absorb before reducing blood glucose.
Analogously, ingested carbohydrates are absorbed over time before increasing blood glucose levels. Time-varying endogenous glucose production, motion, energy expenditure, and natural diurnal variation in insulin sensitivity further complicate dynamics. 

Statistical methods can find patterns in glucose, insulin, and carbohydrate sequences that are predictive of future glucose values. 
However, these detected patterns may not be stable or structured enough to form reliable long term predictions.

On the other hand, physiological models of insulin-glucose dynamics can --- in controlled settings --- describe the evolution of blood glucose farther into the future.  The T1D simulator we study, the UVA/Padova simulator, is a physiologically-grounded model that describes the interaction of glucose, insulin, and orally ingested carbohydrates within different subsystems of a T1D patient \citep{man2014uva}. While the simulator is faithful to T1D physiology, it was not designed to be robust to the noise and missing observations commonly found in CGM, insulin pump, and meal log data. 
Additionally, the UVA/Padova simulator does not account for fluctuations in insulin sensitivity and meal absorption rates, limiting its application to short-range forecasts. 

Here we present a unified model that that fuses two distinct components --- the structured UVA/Padova simulator and a deep state-space model --- that balances the useful inductive bias of the physiological simulator with the flexibility of a modern machine learning sequence model to describe complex dynamics in time series data. 

The UVA/Padova simulator parameters, which represent insulin/carbohydrate absorption rates and sensitivities, undergo physiologically plausible variation over time that we describe with a deep state-space model. 
The result is a parsimonious representation of blood glucose that describes short time scales with the UVA/Padova model (for which it was designed), and long-range temporal variation with the deep state-space model. 
We formalize this fused approach within a single probabilistic generative model, which we call the Deep T1D Simulator (\dtdsim).
First, we describe the full \dtdsim generative model.  We then unpack the individual model components in the following two subsections. 

\begin{figure*}[t!]
\centering
\includegraphics[width=\textwidth]{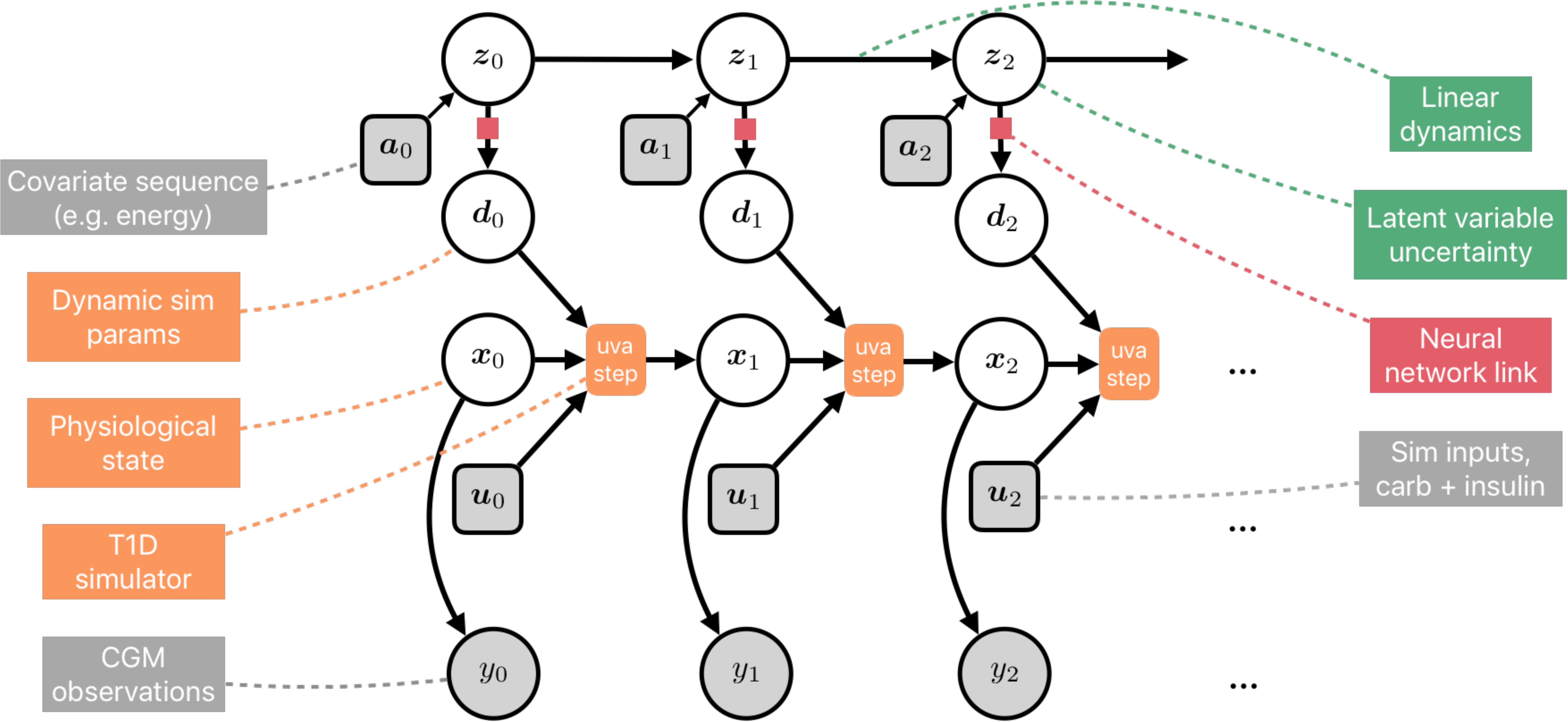}
\caption{Graphical depiction of the \texttt{DTD-sim} model.}
\label{fig:dtd-sim-model}
\end{figure*}

\paragraph{The \dtdsim Model} 
The generative model assumed for the \dtdsim model is depicted graphically in Figure~\ref{fig:dtd-sim-model}.  The aim is to learn a non-linear mapping such that the time-varying parameters to the UVA/Padova simulator can be well-modeled with linear dynamics in some latent space. 
The \dtdsim model is formally specified as:
\begin{align}
\bz_0 &\sim \mathcal{N}(\bmu_0, \bSigma_0) & \text{initial latent state} \\
\bz_t &\sim A \bz_{t-1} + B \ba_t + Q^{1/2} \beps_t, \; \beps_t \sim \mathcal{N}(\vv{0}, I) & \text{latent temporal dynamics} \label{eq:z_dynamics} \\
\bd_t &= \texttt{NN}_{\phi}(\bz_t) & \text{dynamic simulator params.} \\
\bx_t &= \texttt{UVA-step}(\bx_{t-1}, \bd_t, \bu_t, \bs, \Delta_t) & \text{T1D simulator}\\
y_t   &\sim \mathcal{N}(\texttt{CGM}(\bx_{t}, \bs), \sigma^2) & \text{CGM observation}
\end{align}
where $\bz_t \in \mathbb{R}^D$, $\bd_t \in \mathbb{R}^K$, $\bx_t \in \mathbb{R}^{13}$, $\bu_t \in \mathbb{R}^J$, and $\bs \in \mathbb{R}^J$.
The dimensionality of the latent space $D$ can be tuned.  The physiological state $\bx_t$ size is fixed by the simulator definition.  The simulator parameters chosen to be dynamic, $K$, and static, $J$, is a hyperparameter setting, which we fix for this work and describe in Appendix~\ref{sec:uva-padova-sim}. 

\dtdsim incorporates the following modeling components:
\vspace{-.5em}
\begin{itemize} \itemsep 0pt
  \item linear dynamics of the latent state, $\bz_1, \dots, \bz_T$, parameterized by the dynamics matrix $A$, input matrix $B$, process covariance $Q$, and initial mean $\bmu_0$ and covariance $\bSigma_0$,
  \item A non-linear mapping from the latent state $\bz_t$ to the time-varying
    simulator parameters $\bd_t$, modeled as a neural network parameterized
    by $\phi$,
  \item \texttt{UVA-step} integrating the UVA/Padova ODEs, which evolves the
    physiological state $\bx_t$ as a function of patient-specific
    dynamic and static parameters ($\bd_t$ and $\bs$, respectively) and insulin delivery
    and carbohydrates ingested ($\bu_t$) over a period of time $\Delta_t$, and
  \item the observed CGM value $y_t$ at time $t$, modeled via a normal with mean  $\texttt{CGM}(\bx_t, \bs) \triangleq \bx^{(6)} / V_G$, where $V_G$ is a parameter in $\bs$.
\end{itemize}
\vspace{-.5em}
The parameters to be estimated are $\btheta = \{ A, B, Q, \bmu_0, \bSigma_0, \bs, \phi, \sigma \}$; we construct a variational \emph{maximum a posteriori} estimate of $\btheta$.
In the following two sub-sections we describe the T1D simulator and the proposed deep state-space model, both of which present challenges for performing parameter estimation.  We address these challenges in Section~\ref{sec:inference}. 

\subsection{T1D Simulator}
The UVA/Padova T1D simulator represents the instantaneous state of various subsystems of the body, how it changes over time (i.e.~dynamics) and how it is driven by inputs (e.g.~insulin and ingested carbohydrates). 
We denote the instantaneous state at time $t$ as $\bx_t$ for times $t=1, \dots, T$.\footnote{In the exposition, we overload $t$ to represent both index and time value, i.e. $t$ should be thought of as integer values in $[0,T]$. We assume that the time horizon of interest has been rescaled so that the times of measurement are one unit apart.} 
How $\bx_t$ changes over time is defined by instantaneous dynamics 
\begin{align}
  \frac{d \bx_t}{d t} &= f^{(uva)}(\bx_t, \bu_t, \bp) \, ,
\end{align}
where the dynamics are a function of the current state, time-varying inputs, and static parameters, respectively.\footnote{Note that we have not included the \emph{dynamic} parameters $\bd_t$ in the previous section, as the original model only features static subject-specific parameters $\bp$.  How these dynamic parameters factor in will be described in Section~\ref{sec:deep-state-space-model}.}
The evolution from state $\bx_{t-1}$ to $\bx_t$ involves integrating these dynamics over the time increment $\Delta_t$, which in this work we assume is always one so that:
\begin{align}
    \bx_{t+1} &= \int_{t}^{t+1} f^{(uva)}(\bx_{t'}, \bu_t, \bp) dt'  \\
    &\triangleq \texttt{UVA-step}(\bx_{t}, \bu_t, \bp) \, .
\end{align}
Practically, this step integral can be computed using an ODE solver such as Euler's method or Runge-Kutta methods~\citep{burden2015numerical}.

The model represented by $\bx_t$ and $f^{(uva)}$ is a highly constrained yet complex system developed over a series of papers that spans two decades \citep{dalla2002oral, dalla2007meal, dalla2009physical, man2014uva} and is rooted in the seminal work of \citet{bergman1981physiologic}.
The components of the state vector $\bx_t$ correspond to interpretable quantities.  For instance, a set of components of $\bx_t$ describe the \emph{subcutaneous insulin delivery} sub-system, including a dimension that takes delivered insulin as an input. Similarly, other dimensions of $\bx_t$ describe the \emph{oral glucose} sub-system models the process by which ingested carbohydrates become measured glucose, including a dimension that is driven by grams of ingested carbohydrates as an input. 

The nonlinear function $f^{(uva)}(\bx_t, \bu_t, \bp)$ describes the instantaneous change in state $\bx_t$ over time.
These dynamics are driven inputs into the system --- grams of carbohydrates ingested and units of insulin delivered --- represented by $\bu_t$. 
Dynamics are also specified by subject-specific parameters $\bp$, which describe (among other aspects) the rate of absorption of carbohydrates and insulin and the sensitivity of blood glucose to absorbed insulin concentration. 

We implemented the T1D simulator described in \citet{man2014uva} within the automatic differentiation framework JAX~\citep{jax2018github}. We use an Euler integration step to solve the ODE at each time $t$.  See Appendix~\ref{sec:uva-padova-sim} for details on all components of $\bx_t$, simulator parameters, and details of the time-derivative function. 

\begin{figure*}[t!]
\centering
\includegraphics[width=\textwidth]{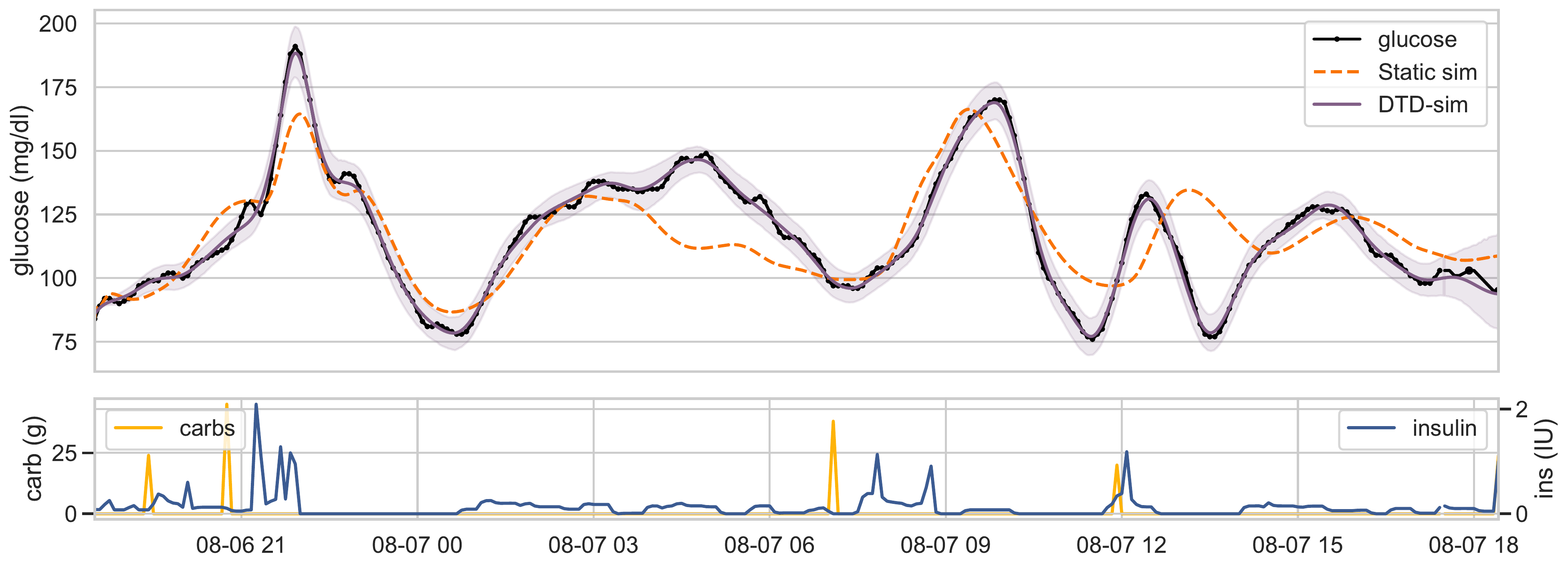}
\caption{\emph{Static} UVA/Padova lacks the capacity to describe real CGM data.  \emph{Top}: model fit comparison of \emph{static} UVA/Padova simulator and \emph{dynamic} \texttt{DTD-sim} model for a full day of CGM data.  \emph{Bottom}: corresponding insulin and carbohydrate data.  Without varying parameters in time and accounting for data noise, the T1D simulator model does not describe observed data.
}
\label{fig:static-uva}
\end{figure*}

The UVA/Padova T1D simulator was designed to validate insulin delivery policies \emph{in silico} over short periods of a few hours at a time --- a task for which it is FDA-approved.  It was not designed, however, to model continuously collected glucose monitor, insulin pump, and carbohydrate log data over the course of weeks and months.  These in-the-wild data are riddled with sources of variability that frustrate the direct application of the UVA/Padova model, stemming from time-varying subject sensitivities, noisy measurements and movement.  To illustrate this point, we compare the \emph{static} UVA/Padova data fit to the \texttt{DTD-sim} data fit, depicted in  Figure~\ref{fig:static-uva}.  
The static model does not have the capacity to describe the variability present in noisy data. 
However, when parameters are allowed to smoothly vary in time, the simulator is able to describe the data over long periods of time.

\subsection{Deep State Space Model}
\label{sec:deep-state-space-model}
To augment the capacity of the T1D simulator model, we allow some of the originally static parameters $\bp$ to vary in time.
We separate the static and time-varying parameters into two vectors denoted $\bs$ and $\bd_t$, where $\bp_t \triangleq (\bd_t, \bs)$ represents the concatenation of static and dynamic parameters. 

We use a state-space model to capture the temporal structure of these time-varying parameters.
A state space model describes the evolution of a stochastic process in terms of transition dynamics.
We use linear Gaussian dynamics to describe the evolution of $\bz_t$
\begin{align}
    \bz_t &= A \bz_{t-1} + B \ba_t + Q^{1/2} \beps_t \, ,
\end{align}
where $\beps_t \sim \mathcal{N}(0,I)$, and $\ba_t$ is a time-varying input sequence of covariates. 
Linear Gaussian dynamics admit computationally tractable inference routines, but assume the restrictive assumption that the latent variable has a multivariate normal distribution \citep{murphy2012machine}.
The parameters $\bp_t = (\bd_t, \bs)$ of the UVA/Padova simulator are a physiological representation of the underlying system and thus must satisfy complex constraints --- to assume $\bd_t$ evolves according to linear dynamics is oversimplified.
To bridge this gap, we learn a neural network to link the latent variable $\bz_t$ to the dynamic parameters $\bd_t$ fed into $\texttt{UVA-Step}$ at each time step. 
We denote the parameters of the neural network link function as $\phi$, $\bd_t = \texttt{NN}_{\phi}(\bz_t)$.
In this work, we use a multi-layer perceptron with two hidden layers of size 128 with rectified linear unit (relu) nonlinearities.

A state-space model is a natural fit to describe the periodic variation typical of a T1D subject.
The well-documented ``dawn phenomenon'' indicates diurnal variation in endogenous glucose production \citep{porcellati2013thirty}.  In fact, recent developments in T1D simulators have begun to incorporate time-variation in insulin sensitivity and endogenous glucose production \citep{visentin2018uva}, albeit with rigidly defined variation over the course of a single day.
Linear Gaussian state space models can describe periodic variation at multiple temporal resolutions and allow the data to dictate the temporal variability of the simulator parameters.

\subsection{Model Fitting and Inference}
\label{sec:inference}
The goal of model fitting and inference is to find a set of parameters $\hat \btheta$ that produces good forecasts.
We use maximum-likelihood estimation to fit $\hat \btheta$, which involves maximizing the marginal log-likelihood of the data $\max_{\btheta} \ln p(\by ; \btheta)$. 
Unfortunately, the introduction of the neural network link function and non-linearities in the \texttt{UVA-step} simulator do not allow for the closed form computation of the marginal likelihood, complicating inference.

To overcome this intractability, we use variational inference methods to optimize a lower bound of the log-marginal-likelihood~\citep{jordan1999introduction, blei2017variational}.
The main issue prohibiting the computation of $\ln p(\by ; \btheta)$ is that the posterior of $\bz_0,\ldots,\bz_T$, $p(\bz_{1:T} | \by ; \btheta)$, cannot be computed in closed form. To circumvent this, variational methods introduce a posterior approximation for the latent variables $\bz$, $q_{\blambda}(\bz_1, ..., \bz_T) = q_{\blambda}(\bz_{1:T})$, specified by variational parameters $\blambda$, resulting in the standard variational objective 
\begin{align}
\mathcal{L}(\btheta, \blambda)
  &= \mathbb{E}_{q_{\blambda} (\bz)}\left[ \ln p(\by \given \bz ; \btheta) \right] - 
     KL(q_{\blambda}(\bz) || p(\bz ; \btheta) ) \label{eq:elbo} \\
  &\leq \ln p(\by \given \btheta)  \, . 
\end{align}
The variational objective is an expectation over the approximate posterior for $\bz_{1:T}$.
Classic variational methods for graphical models \citep{ghahramani2000variational, beal2006variational, blei2006dynamic, wainwright2008graphical} rely on conditionally conjugate structure and closed-form updates to optimize the ELBO.
Due to the non-linear structure in our generative model, we use Monte Carlo estimates of the gradient~\citep{rezende2014stochastic} to maximize Eq.~\eqref{eq:elbo} over $\btheta$ and $\blambda$. 
A prior $p(\btheta)$ can also be incorporated into the variational objective.

\paragraph{Non-centered parameterization} The most common form for the approximate posterior for $\bz_{1:T}$ makes the \textit{mean-field} assumption so that $q_{\blambda}(\bz_{1:T}) = \prod_{t=0}^T q_{\blambda_t}(\bz_t)$, breaking all temporal dependencies \citep{wainwright2008graphical}. 
However, the prior over $\bz_{1:T}$ is auto-correlated by design through Eq.~\eqref{eq:z_dynamics} --- we want the latent process to exhibit smooth dynamics, reflecting the belief that physiological parameters change slowly over time. While algorithmically simple, a mean-field approximation is not appropriate in this scenario.

Instead, we \textit{reparameterize} $q_{\blambda}(\bz_{1:T})$ in terms of the exogenous noise variables $\beps_{1:T}$. Using such an alternative parameterization can be algorithmically beneficial \citep{murray2010slice}.
Instead of approximating the posterior over $\bz_{1:T}$, we can equivalently approximate the posterior over $\beps_{1:T}$ and then deterministically transform this posterior (or its
samples) to obtain an induced approximate posterior over $\bz_{1:T}$.
Because the $\beps_{t}$ are a priori i.i.d.\ standard normal, their posterior can be more accurately modeled with a mean-field approximation.
Additionally, the KL divergence term in Eq.~\eqref{eq:elbo} is simpler to compute.

We approximate the posterior of each $\beps_t$ with a multivariate Gaussian with separate means and covariances (which we take to be diagonal for simplicity) 
\begin{align}
    q_{\blambda_t}(\beps_t) &= \mathcal{N}(\bm_{\blambda}^{(t)}, \mathrm{diag}(\bs_{\blambda}^{(t)})) \, ,
    \label{eq:q_eps}
\end{align}
where $\blambda_t = (\bm_{\blambda}^{(t)}, \bs_{\blambda}^{(t)})$.
Recall that $\bz_t$ is obtained from $\beps_t$ according to
\begin{align}
  \bz_t &= A \bz_{t-1} + B \ba_t + Q^{1/2} \beps_t    \, , \label{eq:q_z_eps}
\end{align}
which shows that $\bz_t$ depends on $\bz_{t-1}$ and the induced posterior of $z_{1:T}$ will capture auto-correlation as desired. 
This dependence is inherited from the structure of the generative
model --- the correlation induced by the dynamics matrix $A$.  The variational parameters
$\bm_{\blambda}^{(t)}$ and $\bs_{\blambda}^{(t)}$ will then alter this
distribution to reflect the information learned from the data.

The variational objective when using the reparameterization in Eqs.~\eqref{eq:q_eps} and \eqref{eq:q_z_eps} is nearly identical to the standard ELBO
\begin{align}
\mathcal{L}(\btheta, \blambda) &=
  \mathbb{E}_{q_{\blambda}(\beps)}\left[ \ln p(\by \given g(\beps), \btheta) + 
    \ln p(\beps \given \btheta) - \ln q_{\blambda}(\beps)\right] \\
  &= \mathbb{E}_{q_{\blambda}(\beps)}\left[\ln p(\by \given \bz, \btheta) \right] 
    - KL\left( q_{\blambda}(\beps) \,||\, \mathcal{N}(0, I) \right) \,. \label{eq:elbo_reparam}
\end{align}
where the expectation is now over $q_{\blambda}(\beps_{1:T})$ and $g(\cdot)$ maps $\beps$ to $\bz$ as defined in Eq.~\eqref{eq:q_z_eps}.
The KL divergence term is a simple analytic function. 
The expected log-likelihood term, however, is still intractable. Because we can easily sample from $q_{\blambda}(\beps_{1:T})$ we instead form a Monte Carlo estimate of the gradient of Eq.~\eqref{eq:elbo_reparam} and use stochastic gradient methods~\citep{rezende2014stochastic}.
Using the whitened space is a simpler alternative to more advanced variational approximations for state space models \citep{archer2015black, bamler2017dynamic} --- incorporation of these techniques may benefit learning $\theta$ in our model setting.

\paragraph{Maintaining stable dynamics} When fitting the dynamics matrix $A$, a practical consideration is ensuring stability in the latent time series.  If the maximal eigenvalue of $A$ is larger than one, $\bz_t$ will diverge over time, leading to unstable forecasts. 
Ensuring the maximal eigenvalue of $A$ is less than one is in tension with the desire to model long term structure in $\bz_t$. If the maximal eigenvalue of $A$ is less than one, the dynamics defined by $A$ will dampen $\bz$, and encourage the process to go to zero when unrolling into the future (i.e.~forming long term forecasts).

To ensure that the eigenvalues of $A$ are close to one in magnitude, we subtract a penalty term from Eq.~\eqref{eq:elbo_reparam} of the form $\mathrm{trace}(A) - D$ --- that is the average eigenvalue should be close to one. Our stochastic gradient updates may make $A$ become unstable, so throughout optimization we project the iterates $A$ back into the set of unit norm matrices by re-normalizing eigenvalues that are greater than one.  We found this projection to be an essential step to reliably learn model parameters $\btheta$ with stochastic gradients.

\subsection{Forecasting}
Given a variational approximation for the latent states up to time $t$, $q_{\blambda}(\beps_{1:t})$, constructing forecasts is a straightforward application of the \dtdsim generative process.   Given data $\by_{1:t}$ and the corresponding variational parameters $\bm_{\blambda}^{(1:t)}$ and $\bs_{\blambda}^{(1:t)}$, we can construct a forecast for a time horizon $h$ by first sampling from the posterior distribution over latent variable dynamics
\begin{align}
  \bz_0 &\sim \mathcal{N}(\hat{\bmu}_0, \hat{\bSigma}_0) \\
  \tilde{\beps}_{1:t} &\sim q_{\blambda}(\beps_{1:t}) && \text{ approx.~posterior sample } \\
  \bz_{t+j} &\sim \hat{A} \bz_{t+j-1} + \hat{B} \ba_{t+j} + \hat{Q}^{1/2} \tilde{\beps}_{t+j}, \; j=1,\ldots,h && \text{ induced posterior }
\end{align}
where $\beps_{s} \sim \mathcal{N}(0, I)$ for $s > t$ and
$\hat{\bmu}_0$, $\hat{\bSigma}_0$, $\hat{A}, \hat{B}$, and $\hat{Q}$ are plug-in estimates of the dynamics
parameters from optimizing the variational objective. 
We then run the simulation forward
\begin{align}
  \hat{\bd}_{1:t+h} &= NN_{\hat{\phi}}(\bz_1), \dots, NN_{\hat{\phi}}(\bz_{t+h}) \\
  \hat{\bx}_{1:t+h} &= \texttt{UVA-solve}(\bd_{1:t+h}, \hat{\bs}, \hat{\bx}_0) \\
  \hat{\by}_{1:t+h} &\sim \mathcal{N}(\texttt{CGM}(\bx_{1:t+h}, \hat{\bs}), \hat{\sigma}^2)
\end{align}
where again $\hat{\bs}$ and $\hat{\sigma}^2$ are plug-in estimates from maximizing Eq.~\eqref{eq:elbo_reparam}.
This procedure produces one posterior predictive sample of $\by_{t+1:t+h}$,
which has a non-Gaussian marginal distribution due to the nonlinearities in the simulation component of the
model.  We use the plug-in Bayes estimate over samples of $\bz$, $\mathbb{E}_{q_{\blambda}(\bz)}\left[
\by_{t+h} \right]$ for forecasts.

%% file: sections/experiments.tex
\section{Empirical Study}
\label{sec:experiments}

Here we describe the empirical study conducted on the cohort detailed in Section~\ref{sec:cohort}. 
We first measure the quality of forecasts produced by our \texttt{DTD-sim} model and compare it to a variety of baseline approaches. 
We then look at generated forecasts under counterfactual future meal and insulin schedules, examining how the different models are influenced by these inputs. 

\subsection{Forecasting glucose at varying time horizons}
\begin{figure*}[t!]
\centering
\includegraphics[width=.49\textwidth]{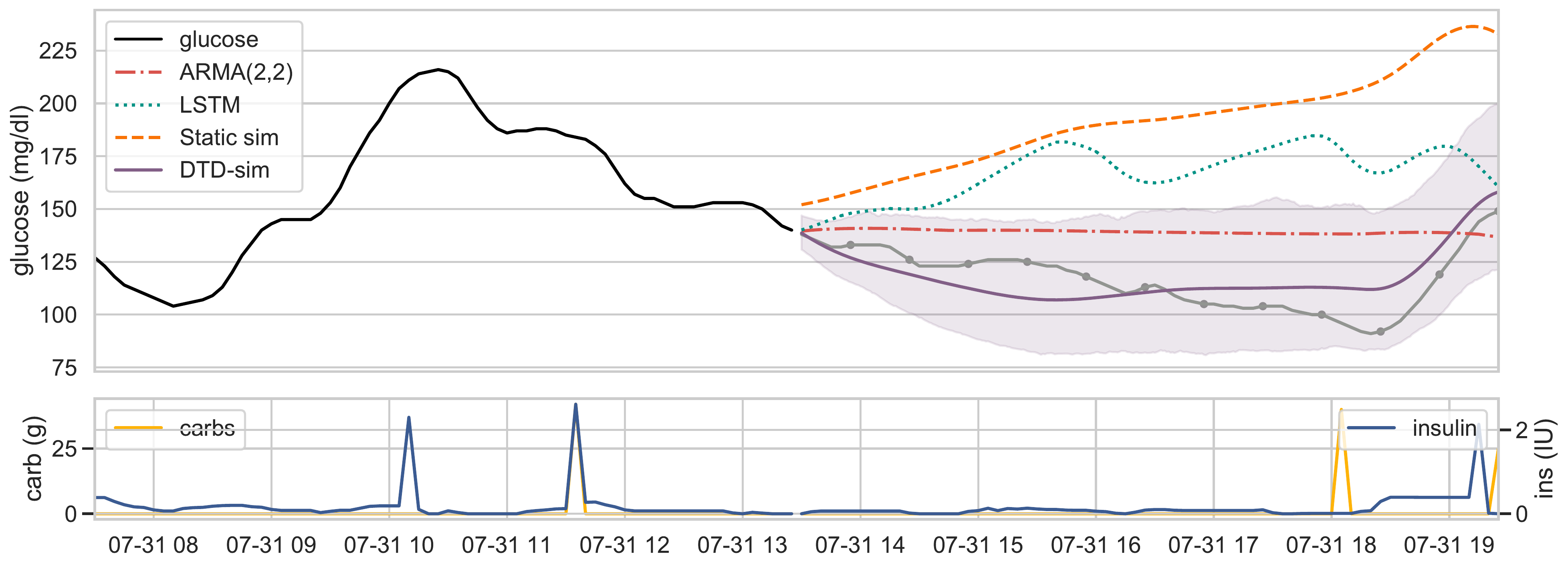} 
\includegraphics[width=.49\textwidth]{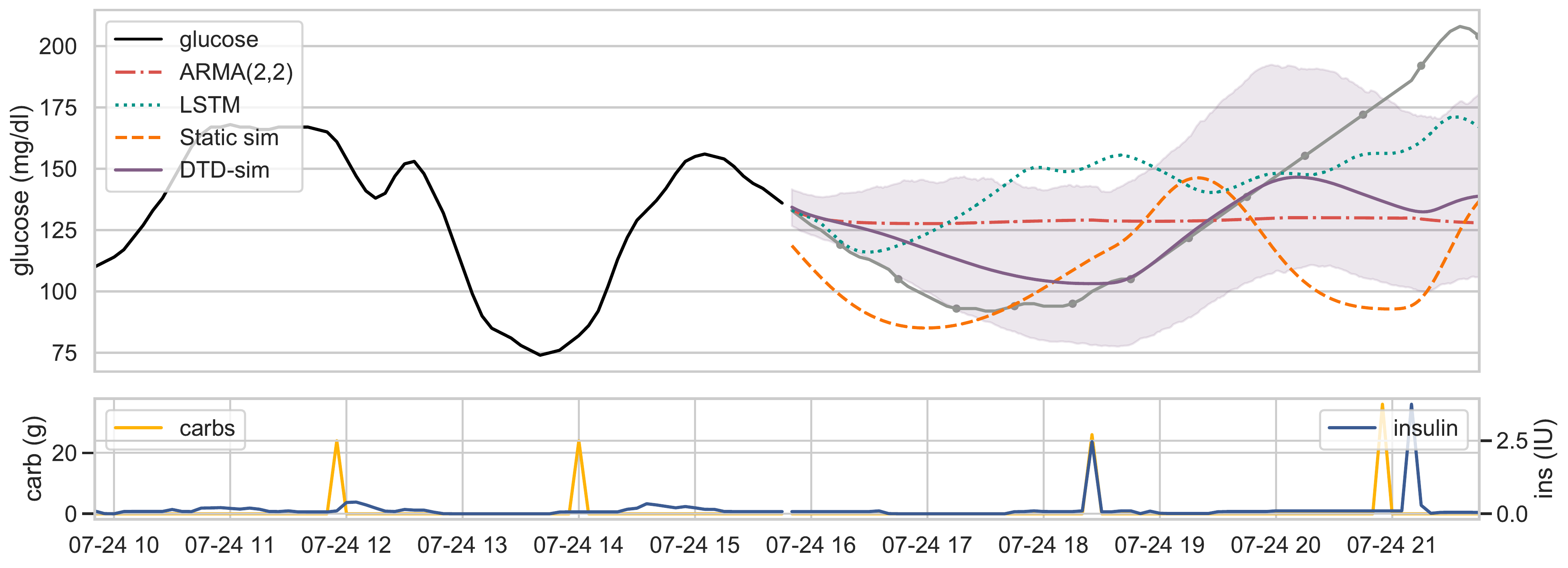} \\
\includegraphics[width=.49\textwidth]{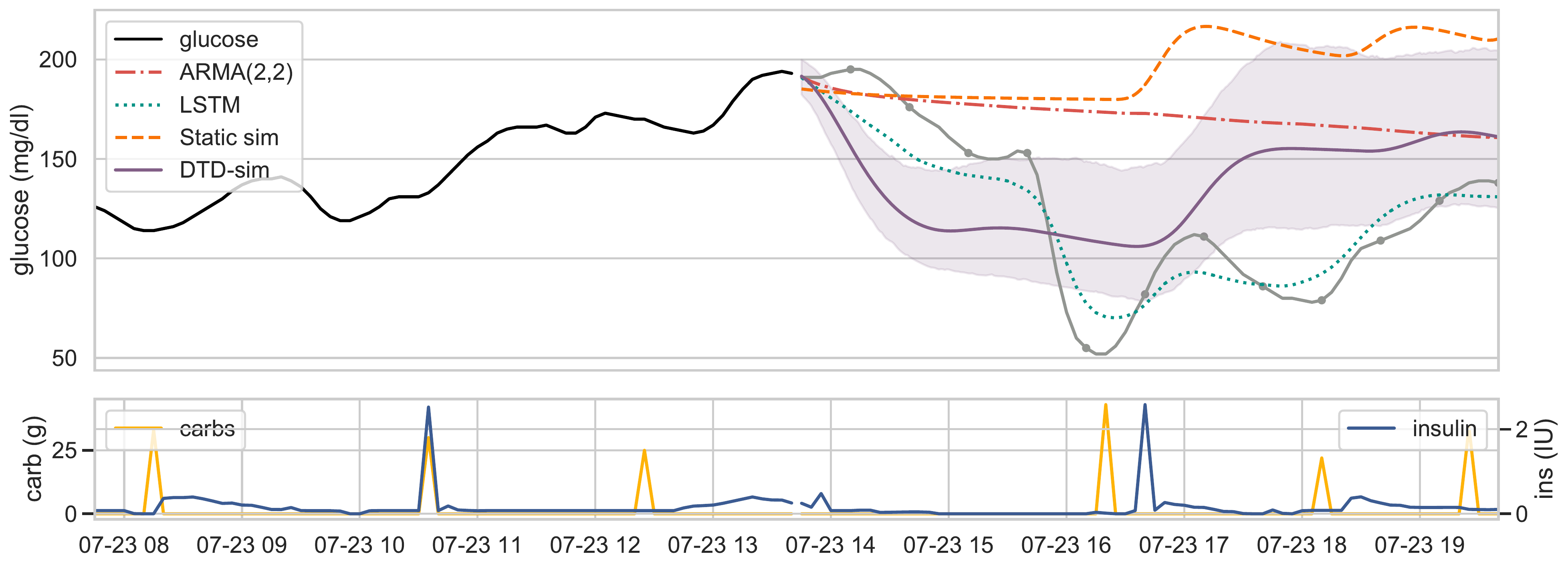}
\includegraphics[width=.49\textwidth]{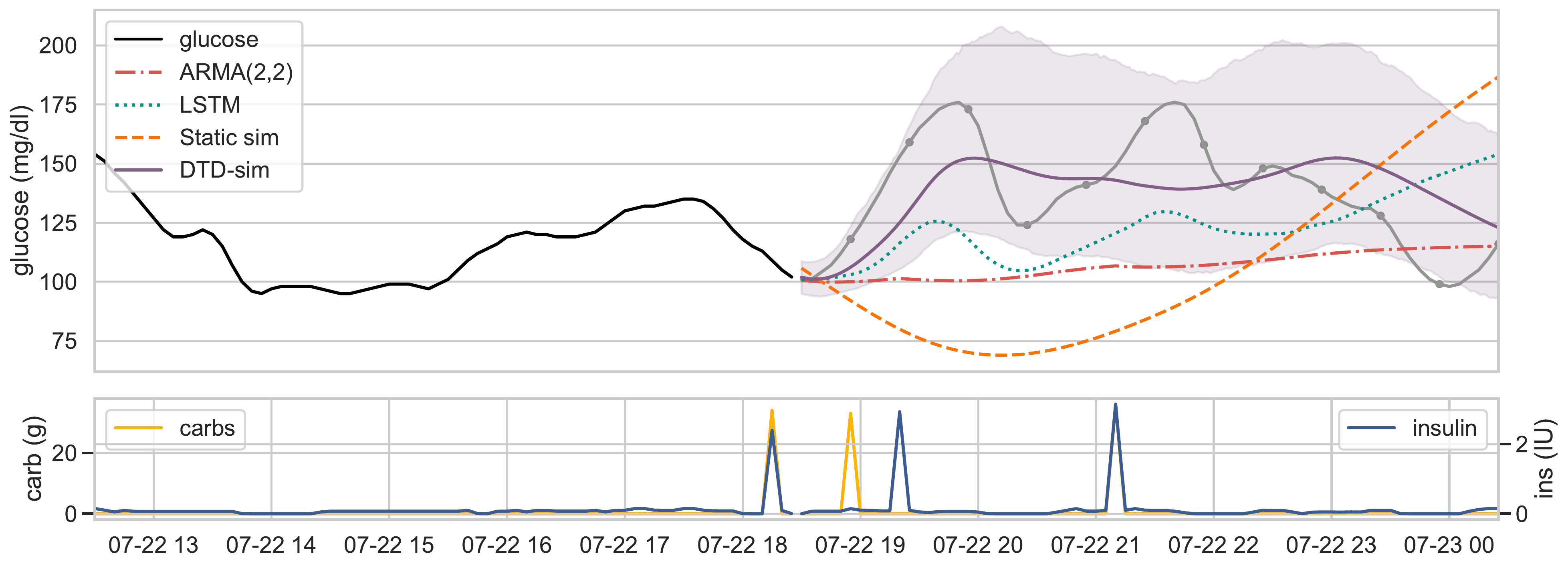}
\caption{A sample of four forecasts from different days.  Observed glucose is in solid black; future glucose is in solid grey with dots indicating every half hour.  In solid purple we depict the proposed method \texttt{DTD-sim} forecasts and 95\% posterior credible intervals.  We also depict the point forecasts for the static UVA/Padova simulator, ARMA(2,2) and LSTM models. }
\label{fig:forecast-comparison}
\end{figure*}

We measure the accuracy of forecasts at multiple time horizons $h$, up to six hours.
For each forecast horizon $h$, we report the mean absolute error (MAE) between the forecast and the true glucose value.
In Appendix~\ref{sec:empirical-supplement} we report additional statistics, including root mean squared error, and mean absolute scaled error (MASE) \citep{hyndman2006another}. 
We also consider predictions in different \emph{contexts}.  
These contexts are defined by time of day, sleep, recent meals, recent bolus injections, 
or elevated/low glucose levels.
Forecast accuracy is more important in some contexts --- for example automated monitoring blood glucose during sleep is crucial for safely avoiding hypoglycemic episodes. 

\begin{figure*}[t]
\floatconts
  {fig:subfigex}
  {\caption{Prediction results by context.  The first context, anytime, is an average over the entire prediction window.  We observe that the \texttt{DTD-Sim} model outperforms both the statistical and mechanistic baselines across all contexts at longer prediction horizons, where $h$ is one to six hours.}}
  {%
    \subfigure[MAE (participant 1)]{\label{fig:forecast-error}%
      \includegraphics[width=.65\linewidth]{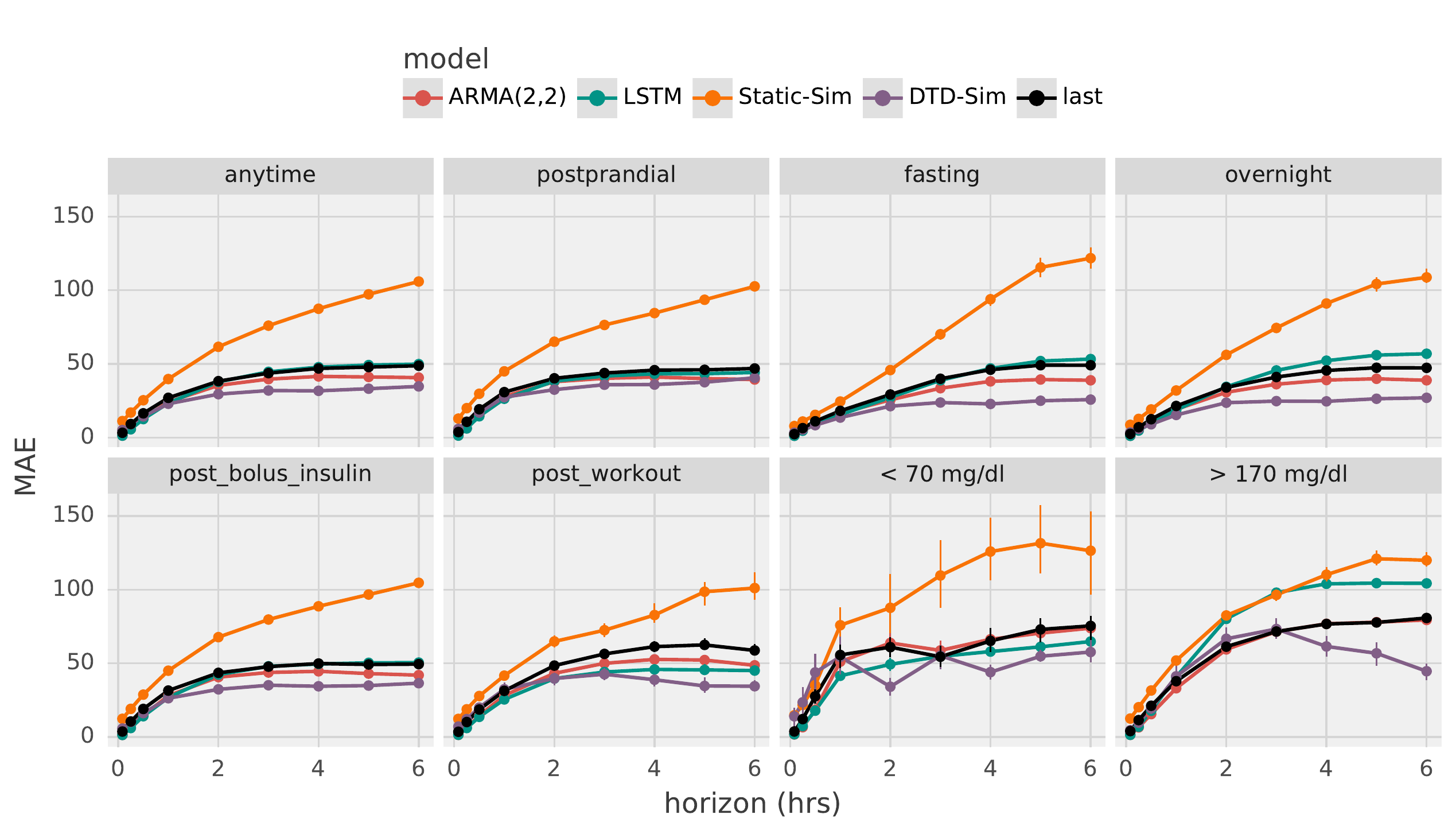}}
    \subfigure[Forecast correlations]{\label{fig:forecast-corr}%
      \includegraphics[width=.3\textwidth]{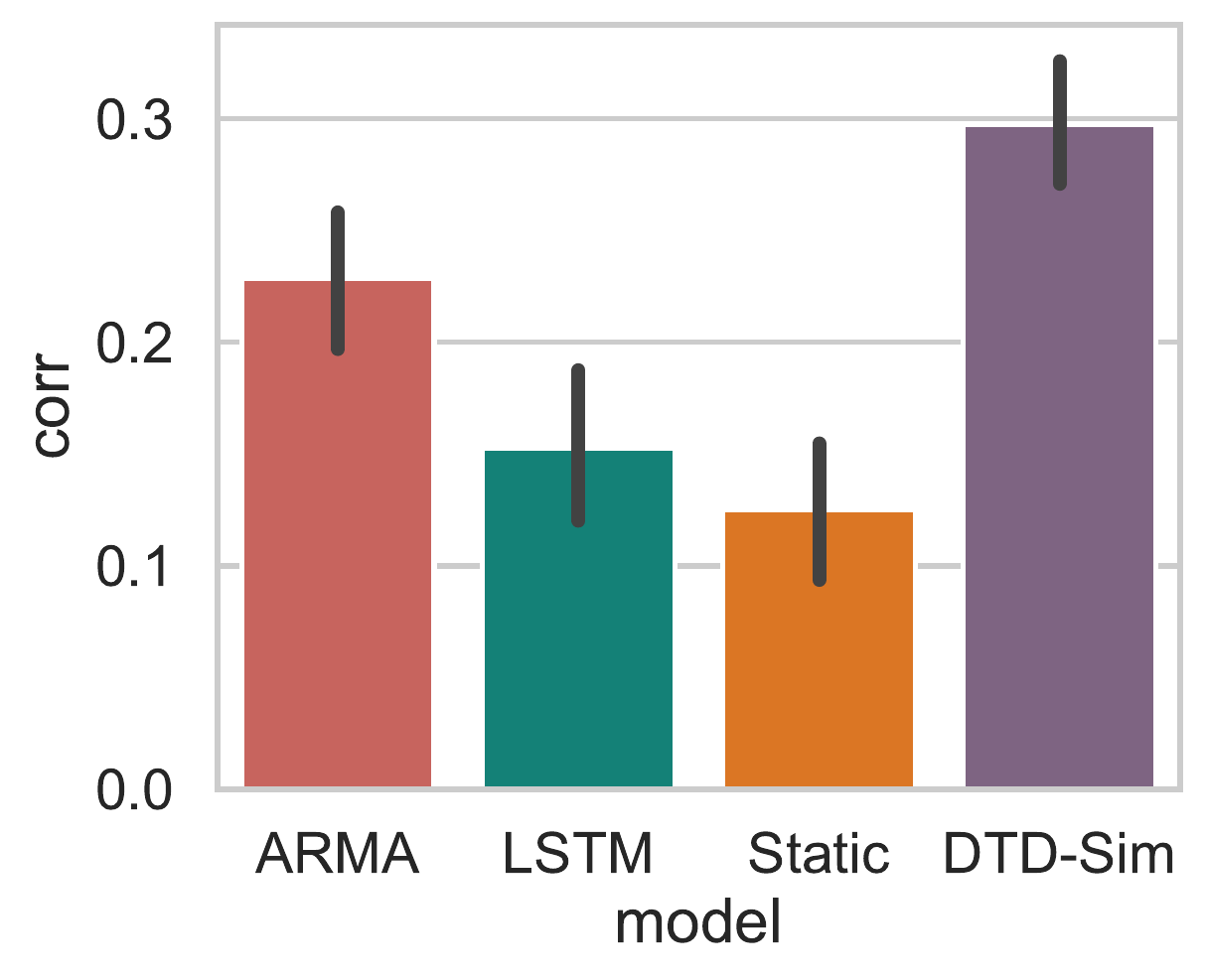}}
  }
  \label{fig:forecast-fig}
\end{figure*}

\paragraph{Baseline methods}
We compare our approach to a handful of baseline approaches for time series forecasting.  
These approaches include purely statistical approaches and a time-invariant simulator.  Specifically, we compare our approach to the following baselines
\vspace{-.5em}
\begin{itemize} \itemsep 0pt
    \item autoregressive moving average (ARMA) models,
    \item long short term memory (LSTM) neural networks,
    \item the static UVA/Padova simulator, and
    \item the last available glucose measurement.
\end{itemize}
\vspace{-.5em}
For all models, we train on the first 90 days, validate using the next 30 days and test on the remaining 31 days (as described in Table~\ref{tab:cohort-summary}).
For the ARMA model, we grid search over both autoregressive order $p$ and moving average order $q$ using a validation set.
The LSTM model includes a forget gate, and hidden states feed into a two layer perceptron with a ReLU nonlinearity; we search over latent state dimension size using a validation set.
Linear time series methods have been used extensively for blood glucose forecasting, and we include the non-linear LSTM model as an additional strong benchmark \citep{montaser2017stochastic, xie2018}.
The static UVA/Padova T1D simulator model is a baseline with fixed simulator parameters over time. 
Because this model cannot describe long periods of time, we re-train the static simulator for each forecast over a running window of data.  Here, we use a moving window of 6 hours to tune model parameters before constructing a forecast. 

We compare these baselines to the \texttt{DTD-Sim} model where we grid search over the latent state dimension $D$ using a validation set. 
The results of the these quantitative experiments are depicted in Figure~\ref{fig:forecast-error}. 
The \texttt{DTD-sim} model performs as well or better than the baselines a few hours after the baseline.  We see a particularly large improvement at much longer horizons --- for example the \texttt{DTD-Sim} model improves upon the baseline LSTM by 5\% at one hour, 27\% at two hours, and 35\% at three hours and 36\% at six hours.
Further, we see improvements at long horizons across most contexts.
The LSTM and ARMA(2,2) models form the best short term forecasts, at 5 to 30 minutes in the future.

The static simulator underperforms the other dynamical models, including the purely statistical models.
This poor performance highlights the unrealistic constraints imposed by the simulator --- that sensitivities and endogenous glucose production are fixed in time.

For a qualitative comparison of model forecasts, we graphically depict a sample of forecast sequences in Figure~\ref{fig:forecast-comparison}.  In these plots we can see some differing behaviors between the models, including the sensitivity to carbohydrate and insulin inputs (which we explore in more detail in the following section). 

While model predictions may be off in terms of mean squared error, the general shape of the forecast can still match the true glucose value quite well.
To quantify this and compare our model, we compute the empirical correlation between the forecast $\hat{y}_{t:t+h}$ and true glucose $y_{t:t+h}$ for $h=6$ hours.
We report the average forecast correlation over $N=1{,}000$ randomly chosen test locations, plotted in Figure~\ref{fig:forecast-corr}.
We observe that, on average, the \texttt{DTD-sim} model consistently produces forecast sequences that correlate more highly than the baseline approaches.

\begin{figure*}[t]
\floatconts
  {fig:subfigex}
  {\caption{The hybrid model balances sensitivity to meals and bolus insulin doses.
  Depicted are four counterfactual scenarios --- bolus/no-bolus insulin and meal/no meal, each one hour after the latest observation.  
  The fully mechanistic static simulator model is overly sensitive to bolus insulin doses and full meals, quickly growing above 200 mg/dl and shrinking to 0 mg/dl.  The AR and LSTM models are less sensitive to bolus doses and full meals.  
  The LSTM model appears to be influenced by a large meal but not a bolus insulin dose.
  The \texttt{DTD-sim} model is both sensitive to bolus insulin and meals, but more stable than the static simulator.
  }}
  {%
    \subfigure[no meal, basal insulin]{\label{fig:synth-lo-lo}%
      \includegraphics[width=.49\textwidth]{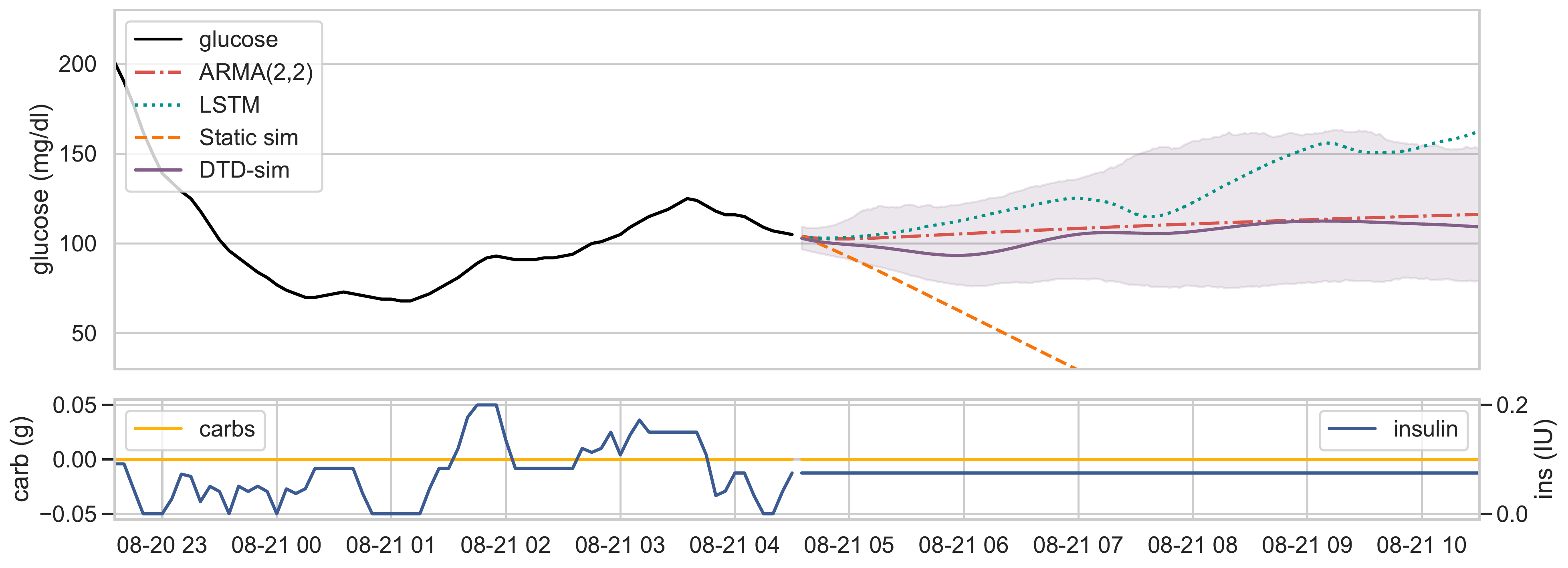}}
    \subfigure[no meal, bolus+basal insulin]{%
      \includegraphics[width=.49\textwidth]{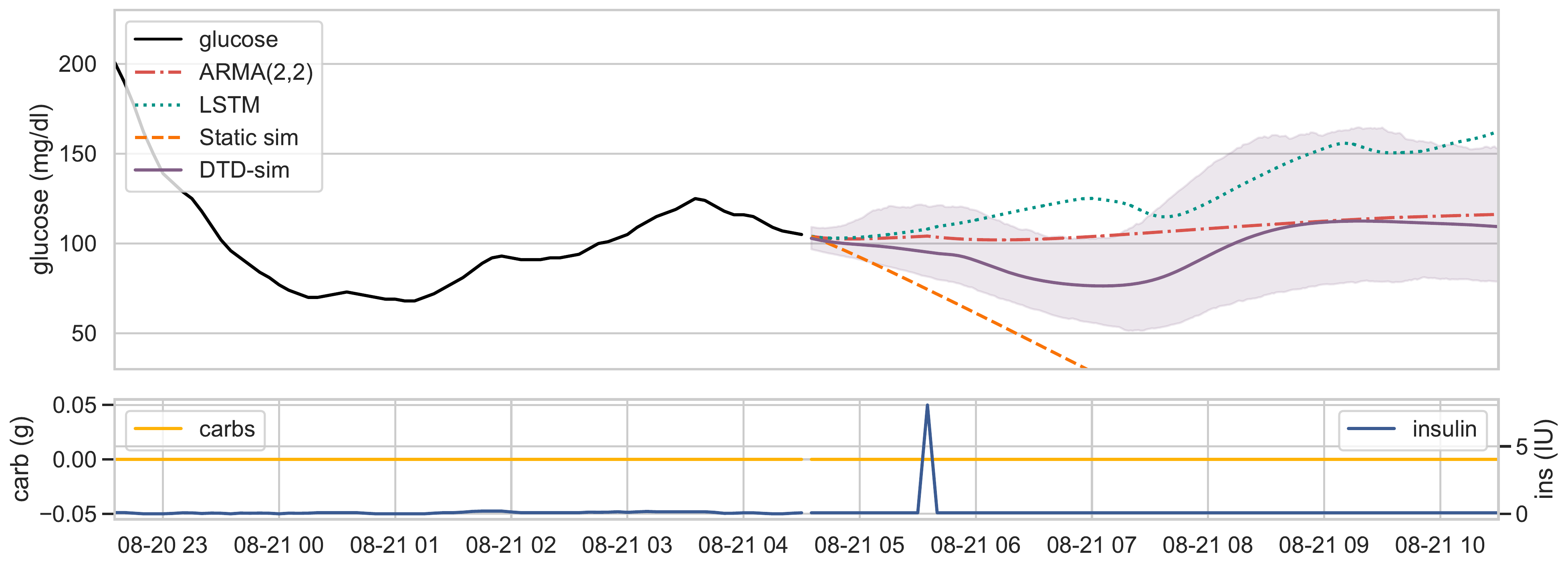}}
    \subfigure[full meal, basal insulin]{%
      \includegraphics[width=.49\textwidth]{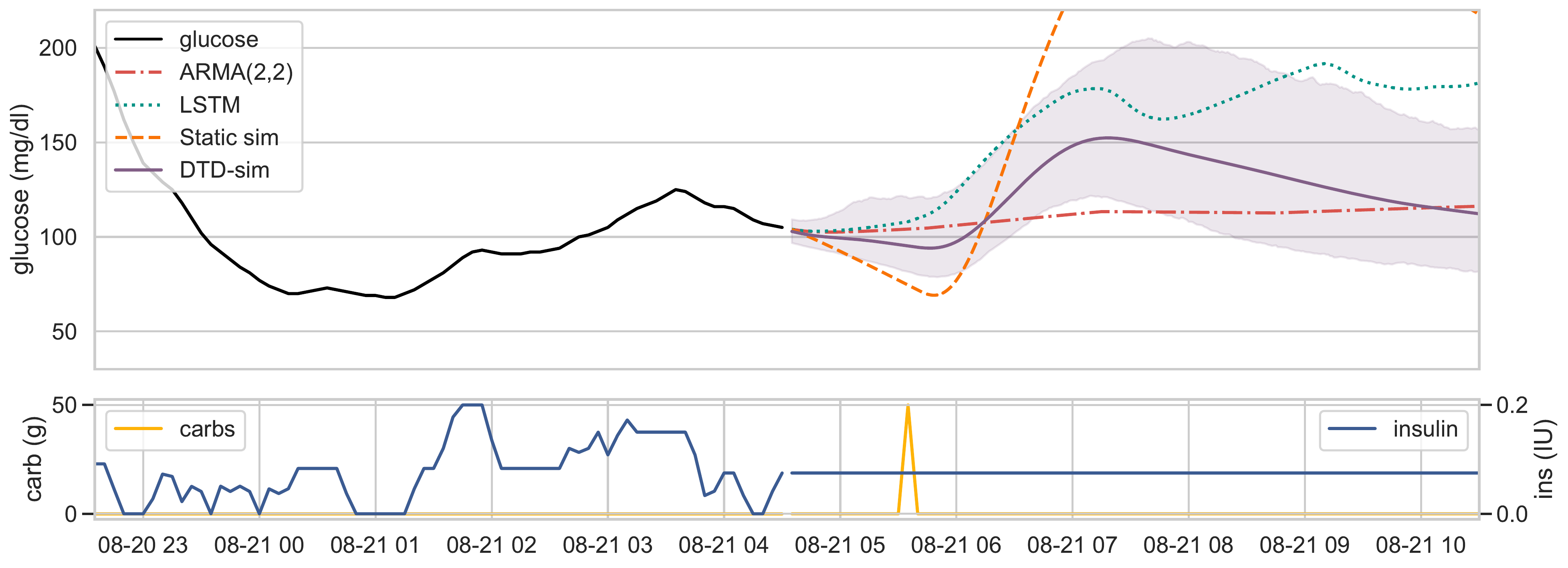}}
    \subfigure[full meal, bolus+basal insulin]{%
      \includegraphics[width=.49\textwidth]{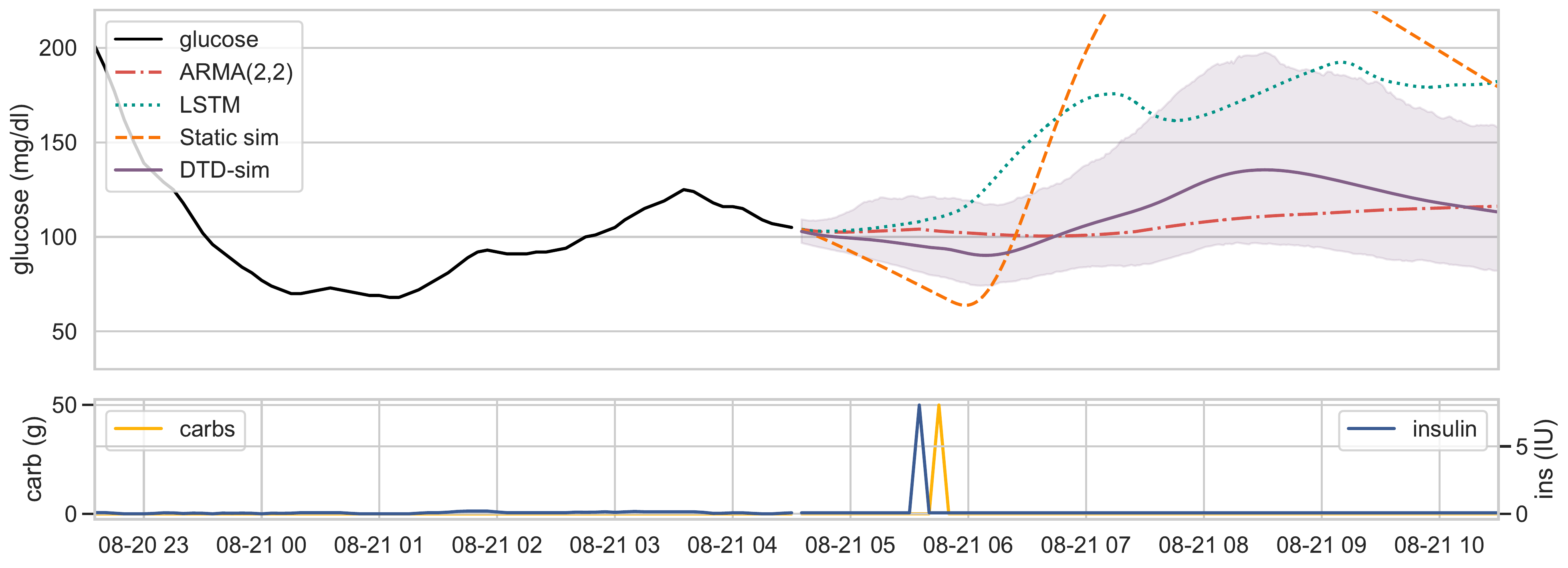}}
  }
  \label{fig:synth-forecasts}
\end{figure*}

\subsection{Counterfactual forecasts}
There is a causal relationship between insulin and carbohydrate inputs and the resulting blood glucose level --- increasing insulin dose should cause the glucose level to fall, and ingesting a large meal should cause the glucose level to rise. 
The T1D simulator encodes this inductive bias in the structure of the differential equation $f^{(uva)}(\cdot)$.
The statistical methods, on the other hand, need to discover this relation from data and could potentially learn a spurious relationship. 

We compare the forecast behavior of the ARMA, LSTM, Static simulator and \texttt{DTD-Sim} models under different counterfactual scenarios. We construct a synthetic insulin delivery and meal schedule and generate forecasts given a fixed sequence of observed data. We consider two settings for each input --- a 50 gram meal compared to no meal and a bolus dose of 8 insulin units compared to no bolus dose (with a constant basal delivery) resulting in four counterfactuals.
We graphically compare these scenarios in Figure~\ref{fig:synth-forecasts}.  

The static simulator is overly sensitive to insulin, going to zero within three to four hours when receiving both a bolus dose or just basal insulin.  Similarly, the static simulator spikes to over three hundred after a meal (with and without the insulin bolus).
The LSTM model appears to be sensitive to carbohydrate inputs, with predicted glucose increasing shortly after the meal.  However, the LSTM model does not appear to be sensitive to an insulin bolus, forming similar forecasts in the bolus and no bolus scenarios. 
The \texttt{DTD-sim} model noticeably reacts to carbohydrate and insulin inputs, with glucose increasing shortly after receiving a meal and decreasing shortly after receiving an insulin bolus. 

Additional synthetic meal and insulin schedule comparisons are depicted in Figure~\ref{fig:synth-forecasts-supp} at randomly selected test points.  We observe a similar pattern of carbohydrate sensitivity and insulin insensitivity for the LSTM compared to the \texttt{DTD-sim} model. 

%% file: sections/discussion.tex
\section{Discussion and Related Work} 
\label{sec:discussion}

\paragraph{Related work} 
As machine learning methods become more pervasive in the sciences, a common goal is to endow ML algorithms with knowledge from applied domains; our work shares this goal with many other approaches.

For example, physics-guided neural networks have been applied in a geological setting \citep{karpatne2017physics}.  This approach introduces constraints on activations and outputs of the neural network model to enforce physical consistency.  In the application of modeling lake temperatures, the physics-guided RNN leverages physical knowledge by pre-training on data simulated from a standard differential equation model of lake temperature \citep{jia2019physics}.
More broadly, simulation models in the physical sciences are an invaluable tool for understanding complex phenomena. 
Machine learning techniques have been used to aid inference in applying simulators to real data \citep{cranmer2020frontier}, develop new models \citep{carleo2019machine}, constrain neural networks \citep{raissi2019physics}, and model control problems \citep{long2018hybridnet}.
With these lines of research we share the common goal of incorporating complex scientific knowledge into a model for real data.
Our approach does not use physical laws or prior knowledge to restrict a flexible model, but rather embeds a physiological model into the data generating procedure.
We introduce the flexibility needed to model real-world observational data by allowing physiological parameters to vary in time according to a sequence model.
A more expansive study of hybrid statistical-and-physical techniques, such as pre-training on simulated data, may yield additional benefits and will be considered in future work. 

Physiological simulators of insulin-glucose dynamics for T1D subjects have been
developed over the past few decades \citep{cobelli2011artificial}.  One line of
research has matured into a FDA-approved simulator
\citep{dalla2002oral,kovatchev2009silico,man2014uva}.  Further enhancements have been considered and tested in lab settings, including improvement of meal
simulation \citep{dalla2007meal} and the incorporation of physical activity
\citep{dalla2009physical}. Physiological models of T1D have been applied to real world CGM and insulin pump data. \citet{liu2019long} demonstrate the utility of a simple physiological model fit using a deconvolution method of the glucose signal. This work, however, does not consider temporal-variation or patterns in subject-specific variables, such as insulin sensitivity. 

The use of tractable latent dynamics with a neural network emission or link function is a common strategy for describing complex observations.  Structured variational autoencoders \citep{johnson2016composing} and deep state-space models \citep{krishnan2017structured} both use variational inference to fit models with complex observations or complex dynamics (or both).
Our approach, while conceptually similar, is distinct in that we model the latent dynamics that capture the variation in simulator \emph{parameters} rather than the data itself.
We model a low-dimensional phenomenon governed by complex, time-varying latent dynamics.

A related line of modeling work incorporates differential equation solvers in probabilistic models \citep{chen2018neural, rubanova2019latent}.  This framework uses neural networks to learn the functional form of the dynamics.
Our goal is to instead make an existing ODE simulator more flexible, but still enjoy the inductive bias described by the simulator. 

\paragraph{Discussion and future work}
Accurately forecasting blood glucose can afford more time to adjust insulin dosage or meals, crucial to the management of T1D. 
To construct more accurate and physiologically plausible forecasts, we integrated a T1D simulator into a machine learning sequence model and applied it to real-world CGM, insulin, and meal log data.
We view the \dtdsim model as a first step in building a reliable glucose forecasting model that will enable better planning and management for T1D. 

We envision many improvements to this model for long-term blood glucose forecasts. 
One obvious shortcoming of our approach is that we are not directly modeling the stochasticity of the input carbohydrates and insulin.  While latent variables can account for some of this uncertainty, a direct model of noise in both the observation of meals and their overall mass could improve forecasts. 

Another direction for improvement is to include additional input sequences as inputs to \dtdsim.  For example, movement, step count, heart rate, or other proxies for energy expenditure may inform the temporal variation in insulin sensitivities.
Explicitly modeling seasonal variation at daily and monthly temporal resolutions may also improve forecast accuracy.
Further, the functional form of $f^{(uva)}(\cdot)$ can likely be improved upon in a data-driven way.  While we assume that $f^{(uva)}(\cdot)$ is fixed, we may want to use the simulator as a starting point and relax the functional form given more observations.  

The expression of T1D varies from person to person.  Our algorithm development has been limited to a small cohort.  With an increased diversity in insulin-glucose observations, a joint model of many subjects through a hierarchy may help improve long term forecasts.

Finally, our hybrid statistical-and-physiological modeling approach may be suitable for adapting other biophysical ODEs that describe complex phenomena over time to model real-world data. 
Models of the cardiovascular system, biomechanics, or long term patient trajectories could be augmented in a way similar to our approach.

%% file: sections/uva-simulator.tex
\section{UVA/Padova T1D Simulator}
\label{sec:uva-padova-sim}
As noted in the main text, the UVA/Padova T1D simulator model represents the instantaneous state of various subsystems of the body as a state vector and dynamics function
\begin{align}
    \frac{d \bx}{dt} &= f^{(uva)}(\bx_t, \bu_t, \bs) \equiv \dot{\bx}
\end{align}
where the components of $\bx_t \in \mathbb{R}^{13}$ represents the instantaneous physiological state of the body at time $t$, $\bu_t \in \mathbb{R}^2$ denote the insulin units and carbohydrate mass delivered at time $t$, and $\bs$ are subject-specific simulator parameters that represent endogenous glucose production rates and insulin sensitivity. 

In the UVA/Padova model, the state  is a thirteen-dimensional vector representing various subsystems.
The oral glucose subsystem contains components
\begin{align}
    \bx^{(1)} &= Q_{sto1} & \text{ first stomach compartment }\\
    \bx^{(2)} &= Q_{sto2} & \text{ second stomach compartment }\\
    \bx^{(3)} &= Q_{gut} & \text{ first stomach compartment }
\end{align}

The glucose subsystem describes two compartment glucose kinetics
\begin{align}
    \bx^{(4)} &= G_p & \text{ plasma glucose } \\
    \bx^{(5)} &= G_t & \text{ tissue glucose } \\
    \bx^{(6)} &= G_s & \text{ subcutaneous glucose (CGM) } 
\end{align}

The insulin subsystem describes insulin kinetics, including the absorption into active insulin
\begin{align}
    \bx^{(7)} &= I_p & \text{ plasma insulin } \\
    \bx^{(8)} &= I_l & \text{ liver insulin } \\
    \bx^{(9)} &= X_L & \text{  } \\
    \bx^{(10)} &= X & \text{ active insulin } \\
    \bx^{(11)} &= \tilde{I} & \text{  } 
\end{align}

Finally, the subcutaneous insulin subsystem describes the absorption kinetics of delivered insulin
\begin{align}
    \bx^{(12)} &= I_{sc1} & \text{ subcutaneous compartment one } \\
    \bx^{(13)} &= I_{sc2} & \text{ subcutaneous compartment two } 
\end{align}

Given the definition of the $\bx_t$ state components, we define the dynamics of each subsystem (along with useful intermediate quantities). 

The oral glucose subsystem evolves as
\begin{align}
    \dot{Q}_{sto1}(t) &= -k_{gri}\cdot Q_{sto1}(t) + D \cdot \delta(t) \\
    \dot{Q}_{sto2}(t) &= -k_{empt}(Q_{sto}) \cdot Q_{sto2}(t) + k_{gri}\cdot Q_{sto1}(t) \\
    \dot{Q}_{gut}(t) &= -k_{abs} \cdot Q_{gut}(t) + k_{empt}(Q_{sto}) \cdot Q_{sto2}(t) \\
    Q_{sto}(t) &\triangleq Q_{sto1}(t) + Q_{sto2}(t) \\
    Ra(t) &\triangleq \frac{f \cdot k_{abs} \cdot Q_{gut}(t)}{BW} & \text{ glucose rate of appearance }
\end{align}

Glucose kinetics are 
\begin{align}
    \dot{G}_p(t) &= EGP(t) + Ra(t) - U_{ii}(t) - E(t) - k_1 \cdot G_p(t) + k_2 \cdot G_t(t) \\
    \dot{G}_t(t) &= -U_{id}(t) + k_1 \cdot G_p(t) - k_2 \cdot G_t(t) \\
    \dot{G}_s(t) &= -\frac{1}{T_s} \cdot G_s(t) + \frac{1}{T_s} \cdot G(t) \\
    G(t) &\triangleq G_p(t) / V_G 
\end{align}
where the endogenous glucose production and insulin-based utilization are defined
\begin{align}
    EGP(t) &= k_{p1} - k_{p2} \cdot G_p(t) - k_{p3} \cdot X_L(t) \\
    U_id(t) &= \frac{ \left(V_{m0} + V_{mx} \cdot X(t) \cdot(1 + r_1 \cdot risk) \right) \cdot G_t(t) }{K_{m0} + G_t(t)} 
\end{align}

Insulin kinetics are defined
\begin{align}
    \dot{I}_p(t) &= -(m_2 + m_4) \cdot I_p(t) + m_1 \cdot I_l(t) + R_{ai}(t) \\
    \dot{I}_l(t) &= -(m_1 + m_3) \cdot I_l(t) + m_2 \cdot I_p(t) \\
    \dot{X}_L(t) &= -k_i \cdot \left( X_L(t) - \tilde{I}(t) \right) \\
    \dot{\tilde{I}}(t) &= -k_i \cdot \left( \tilde{I}(t) - I(t) \right) \\
    \dot{X}(t) &= -p_{2U} \cdot X(t) + p_{2U} \left( I(t) - I_b \right) \\
    I(t) &= I_p(t) / V_I
\end{align}

The subject-specific simulator parameters $\bs$ include 
\begin{align}
    \bs = &(k_{min}, k_{max}, k_{abs}, f, b, d, V_G, k_{1:2}, V_I, m_{1:4}, k_{p1:3}, k_i, \\
          &~F_{snc}, V_{m0}, K_{m0}, I_b, k_{e1:2}, k_{a1:2}, k_d, k_{sc}, BW)
\end{align}
We succinctly express these equations as $f^{(uva)}(\bx, \bu, \bs) : \mathbb{R}^{13} \times \mathbb{R}^2 \times \mathbb{R}^{29} \rightarrow \mathbb{R}^{13}$. 
We augment the existing simulator model by allowing some components of the parameter vector $\bs$ to vary over time.

%% file: sections/app-experiments.tex
\section{Empirical Study Supplement}
\label{sec:empirical-supplement}

Here we include additional plots to support the empirical study. 
The mean absolute scaled error (MASE) is defined as
\begin{align}
    MAE_0(h)  &= \frac{1}{T-h} \sum_{t=h}^T |Y_{t} - Y_{t-h}| \\
    q_t(h)  &= \frac{Y_t - \hat{Y}_{t | t-h}}{MAE(h)} \\
    MASE(h) &= mean(|q_t(h)|) \, ,
  \end{align}
where $MAE_0(h)$ is the error of a simple baseline --- the forecast value
is equal to the latest observation.  Intuitively, MASE measures the MAE ratio between this simple baseline and the predicted model --- a value of 1 indicates no improvement over the baseline.  See~\citet{hyndman2006another} for more details.

\begin{figure*}[t]
\floatconts
  {fig:subfigex}
  {\caption{Prediction results by context.  The first context, anytime, is an average over the entire prediction window.  We observe that the \texttt{DTD-Sim} model outperforms both the statistical and mechanistic baselines across all contexts at longer prediction horizons, where $h$ is one to six hours.}}
  {%
    \subfigure[RMSE (participant 1)]{%
      \includegraphics[width=.8\linewidth]{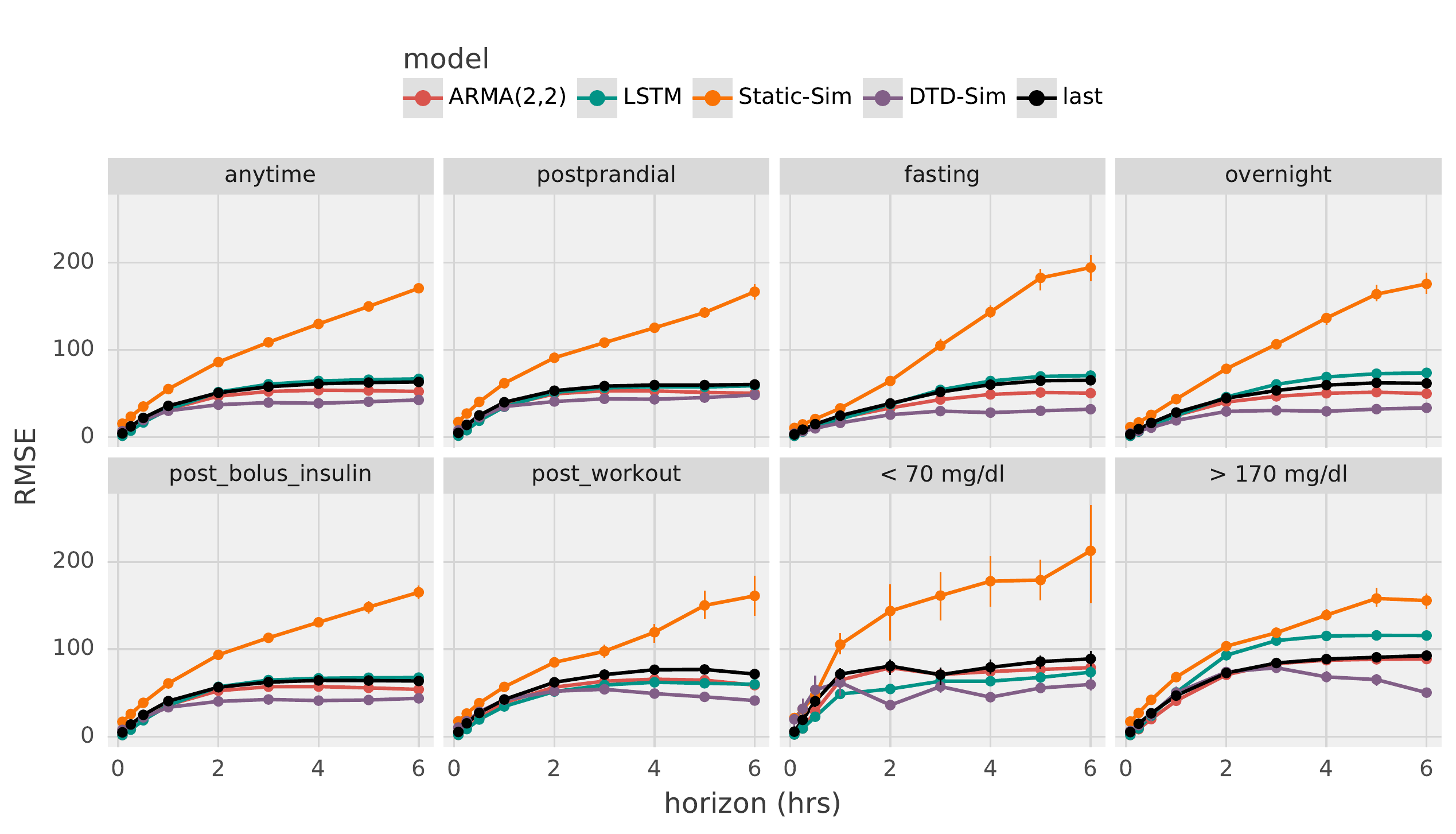}}
    \subfigure[MASE (participant 1)]{%
      \includegraphics[width=.8\linewidth]{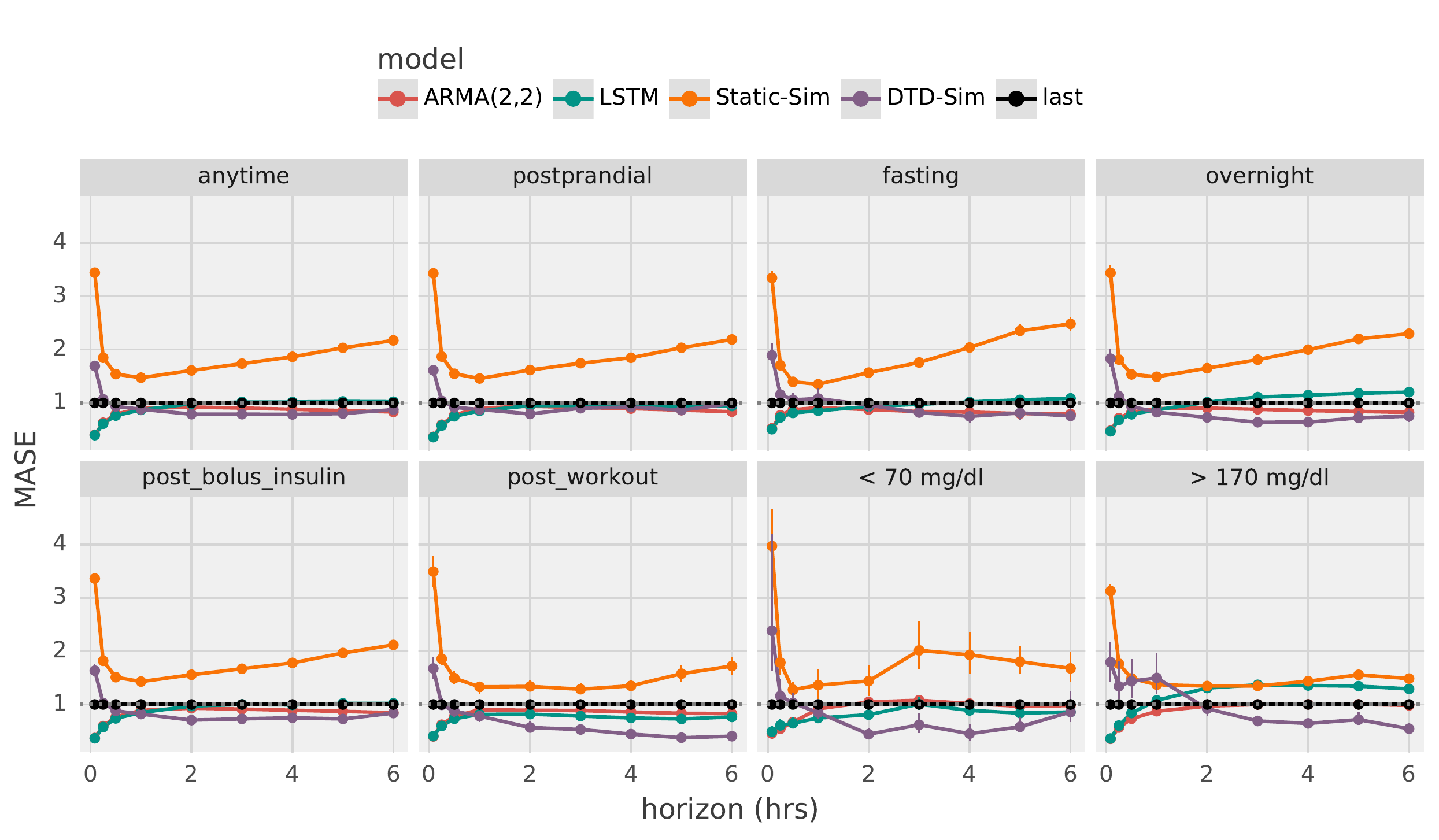}}
  }
  \label{fig:forecast-error-supp}
\end{figure*}

\begin{figure*}[t]
\floatconts
  {fig:subfigex}
  {\caption{Additional synthetic meal comparisons. 
  }}
  {%
    \subfigure[no meal, basal insulin]{%
      \includegraphics[width=.49\textwidth]{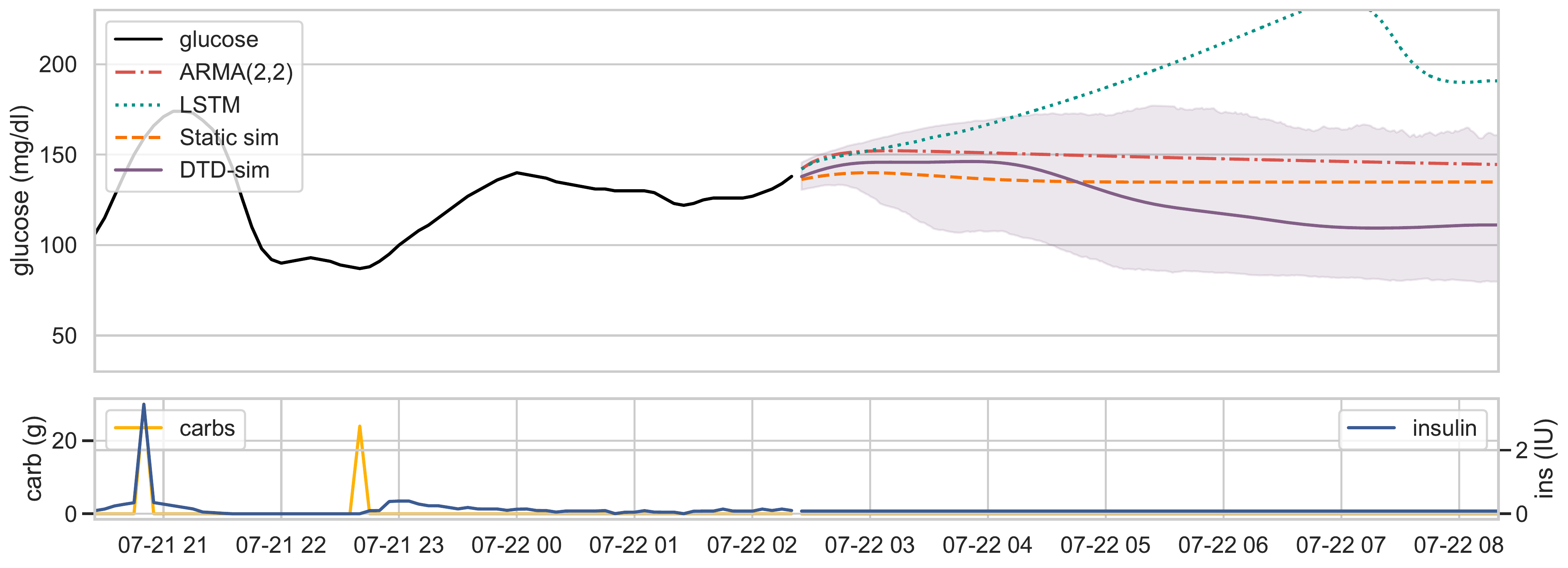}}
    \subfigure[no meal, bolus+basal insulin]{%
      \includegraphics[width=.49\textwidth]{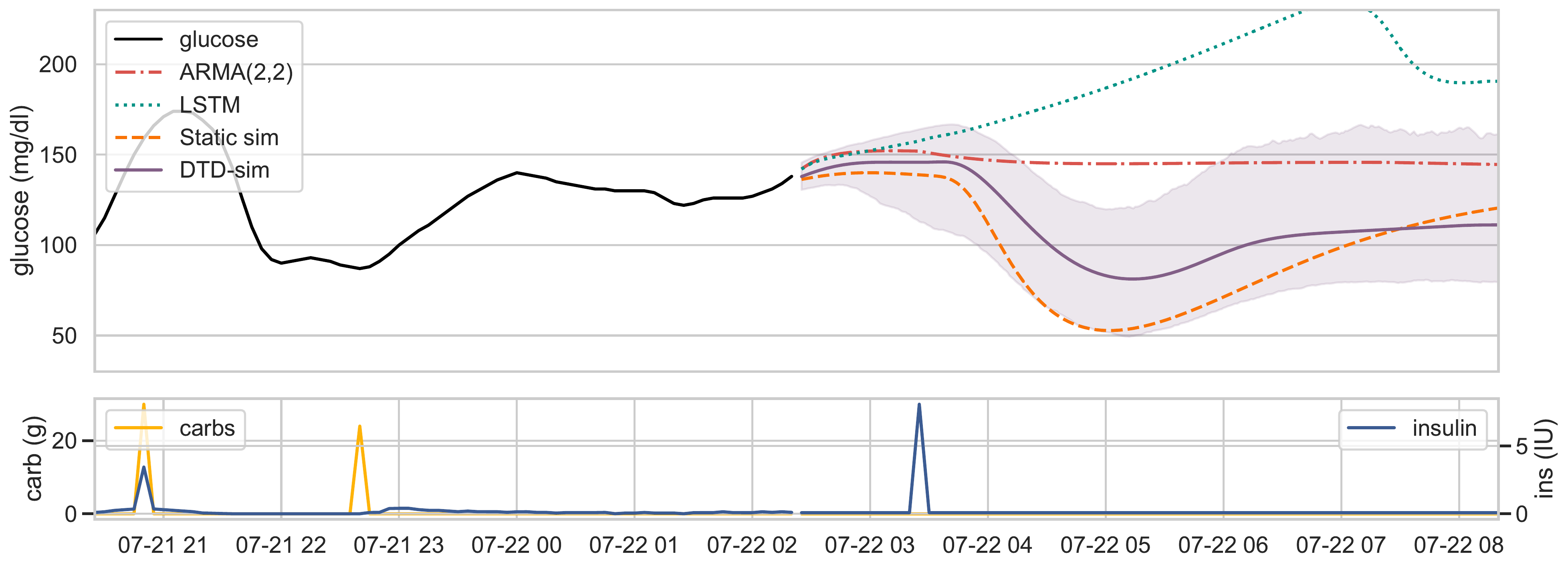}}
    \subfigure[full meal, basal insulin]{%
      \includegraphics[width=.49\textwidth]{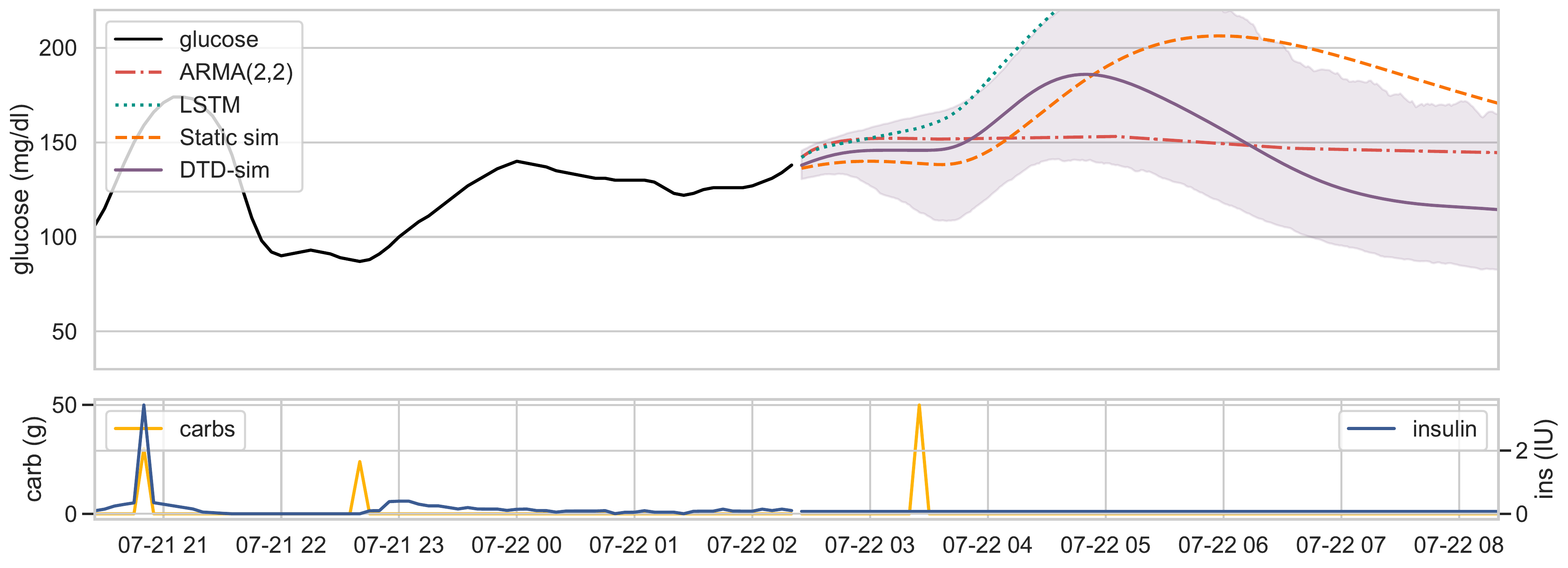}}
    \subfigure[full meal, bolus+basal insulin]{%
      \includegraphics[width=.49\textwidth]{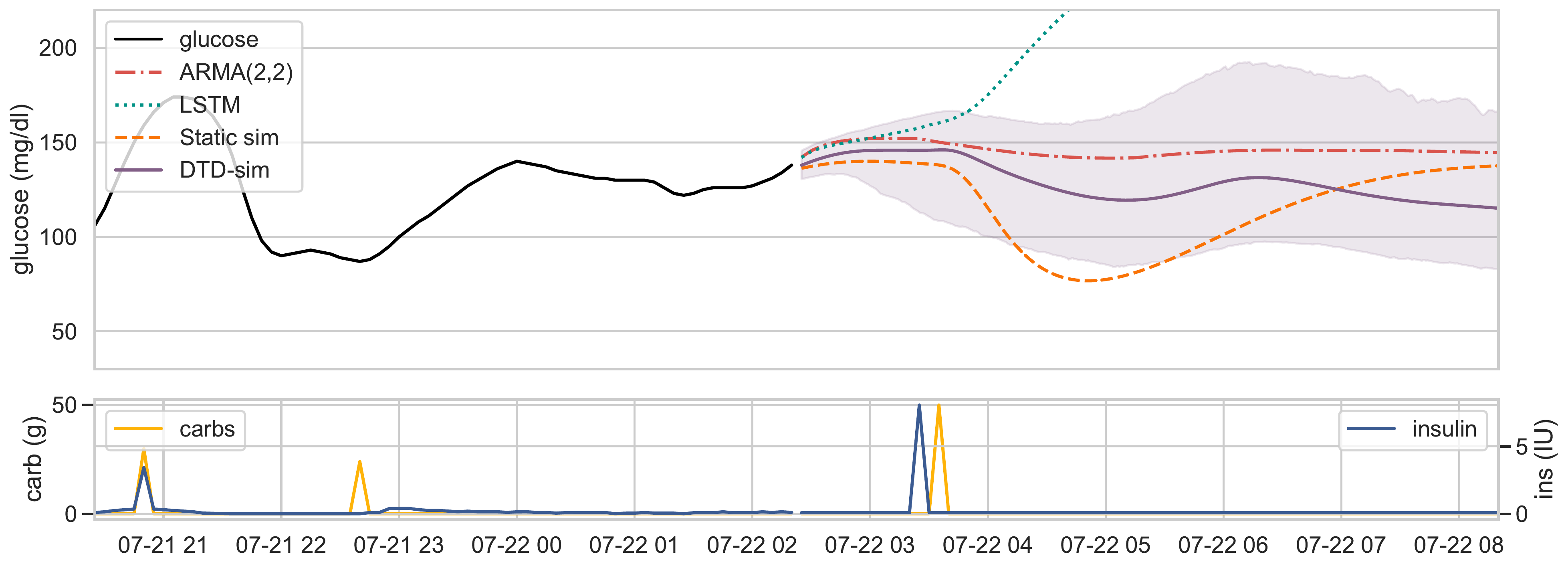}}
  } 
  \label{fig:synth-forecasts-supp}
\end{figure*}

\begin{figure*}[t]
\floatconts
  {fig:subfigex}
  {\caption{Additional synthetic meal comparisons. 
  }}
  {%
    \subfigure[no meal, basal insulin]{%
      \includegraphics[width=.49\textwidth]{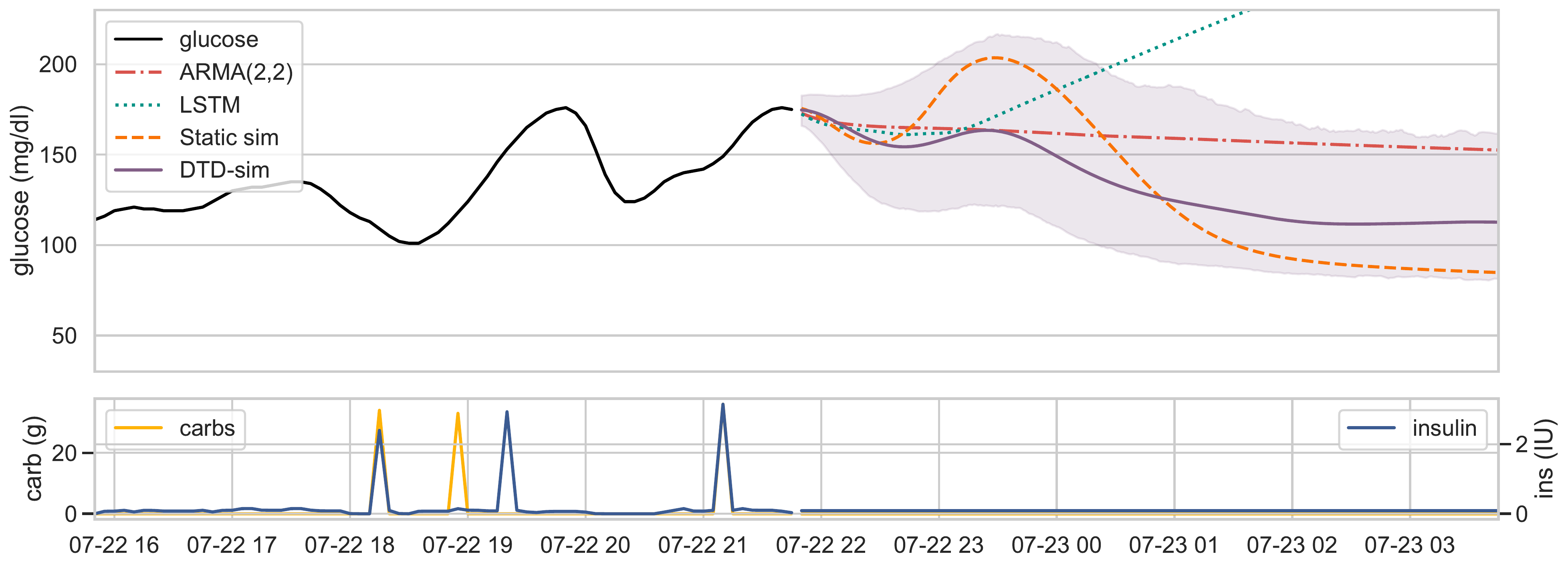}}
    \subfigure[no meal, bolus+basal insulin]{%
      \includegraphics[width=.49\textwidth]{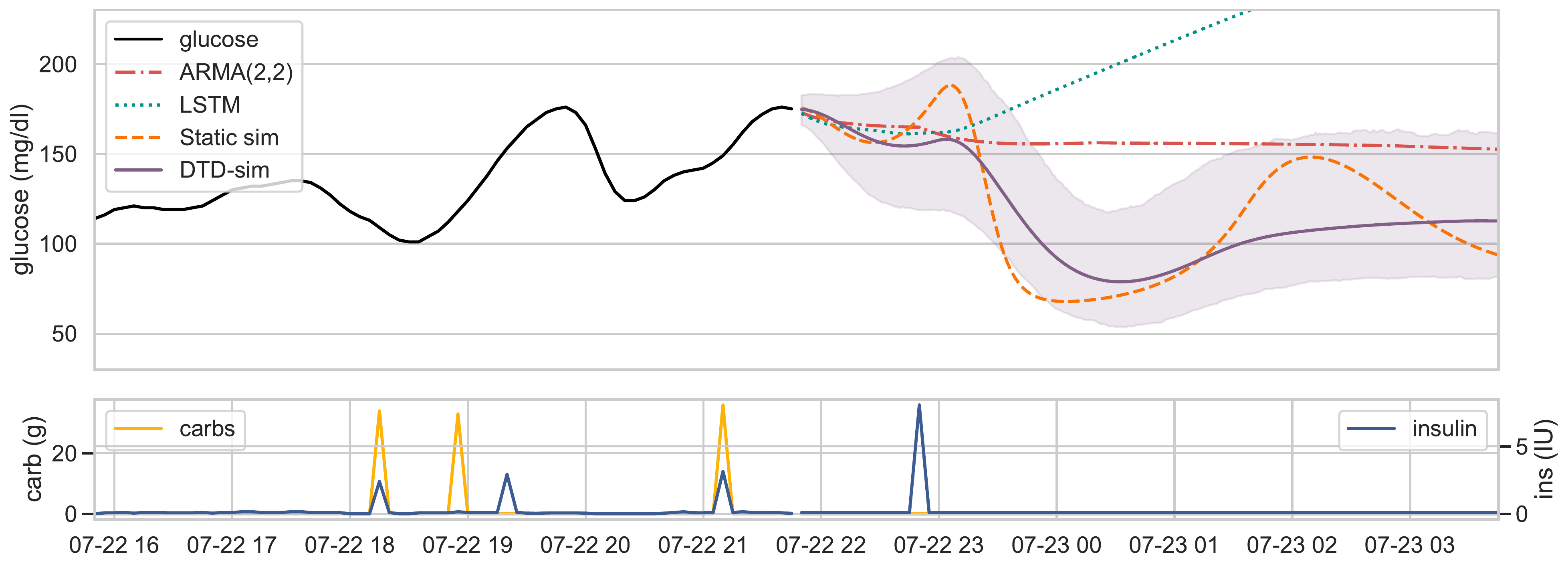}}
    \subfigure[full meal, basal insulin]{%
      \includegraphics[width=.49\textwidth]{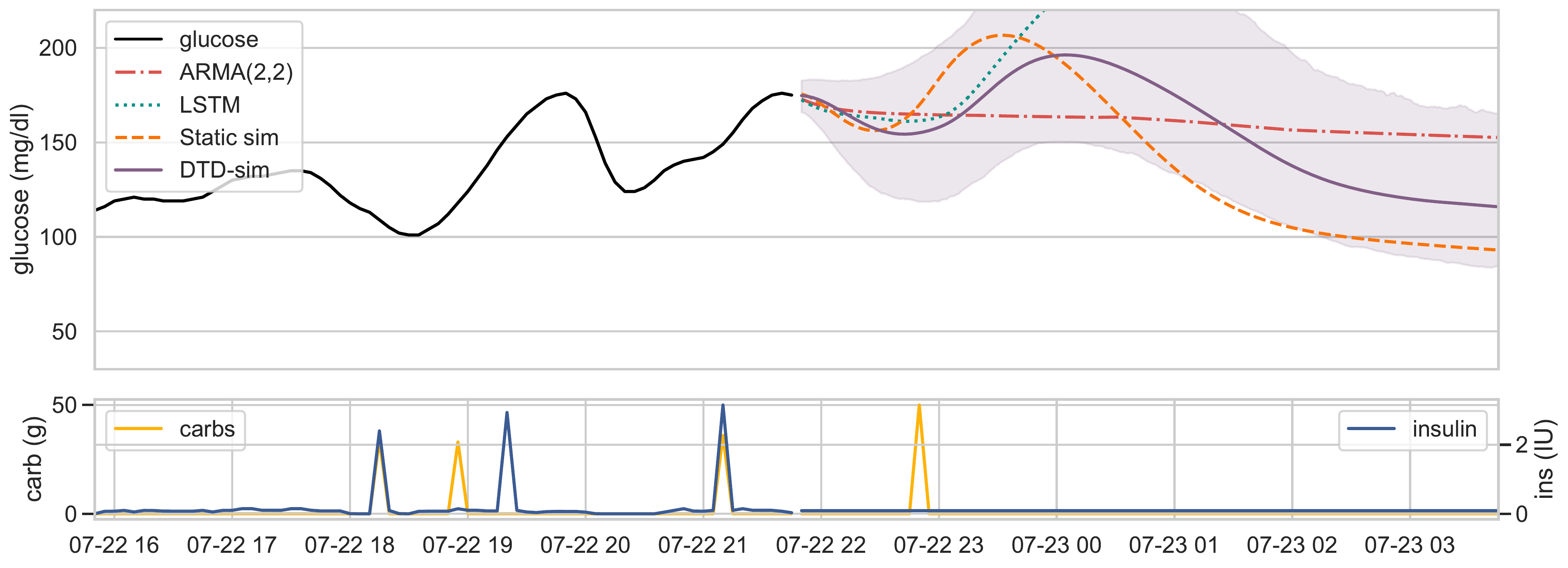}}
    \subfigure[full meal, bolus+basal insulin]{%
      \includegraphics[width=.49\textwidth]{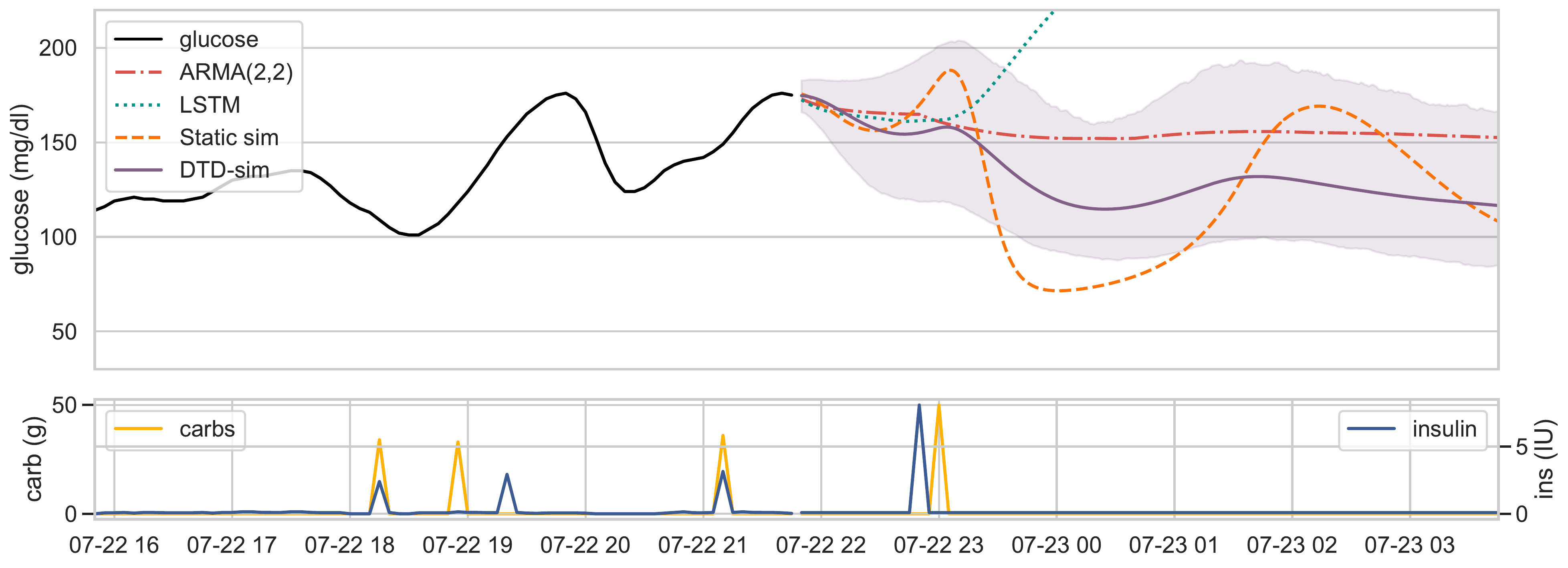}}
  }
  \label{fig:synth-forecasts-supp-b}
\end{figure*}

%% file: sections/app-inference.tex
\section{Inference Algorithm Details}
\label{sec:inference-supplement}

In this appendix we provide a step-by-step overview of our proposed variational inference algorithm for \texttt{DTD-Sim}. Throughout, we use the notation $\mathbf{z}_{1:T}$ to indicate the set of all $\mathbf{z}_t$, $t=1,\ldots,T$, and similarly for other variables.

Each iteration of the algorithm proceeds by sampling the latent trajectories $\mathbf{z}_{1:T}$ and transforming them into the parameters of the UVA-Padova simulator. The interpretable state-space parameter at time $t$, $\mathbf{x}_t$, is then determined by integrating the UVA-Padova system forward one time-step. Then, $\mathbf{x}_t$ is used to compute the expected log-likelihood of the observed data which we differentiate through in order to update the parameters. The detailed steps of the algorithm are as follows:
\begin{enumerate}
    \item Sample $\tilde{\mathbf{\epsilon_t}} \sim \mathcal{N}(\mathbf{0}, \mathbf{I}), t=1,\ldots,T$.
    \item Sample $\mathbf{z}_0 \sim \mathcal{N}(\mathbf{\mu}_0, \mathbf{\Sigma}_0)$.
    \item For $t = 1,\ldots,T$
    \begin{enumerate}
    \item Construct $\mathbf{z}_t = A\mathbf{z}_t + B\mathbf{u}_t + Q^{1/2}\tilde{\mathbf{\epsilon}}_t$.
    \item Compute $\mathbf{d}_t = \mathrm{NN}_{\phi}(\mathbf{z}_t)$.
    \item Compute $\mathbf{x}_t = \texttt{UVA-step}(\mathbf{x}_{t-1}, \mathbf{d}_t, \mathbf{u}_t, \mathbf{s}, \Delta_t)$ using a numerical integration scheme.
    \end{enumerate}
    \item Compute $\nabla_{\mathbf{\lambda},\mathbf{\theta}}\ln p(\mathbf{y}_{1:T} | \mathbf{z}_{1:T} ; \mathbf{\theta})$, where
    \begin{equation}
    \ln p(\mathbf{y}_{1:T} | \bz_{1:T} ; \btheta) = \prod_{i=1}^T \ln p(y_t | \bz_t ; \btheta).
    \end{equation}
    \item Update $\mathbf{\lambda}$ and $\mathbf{\theta}$ according to this gradients.
\end{enumerate}

We used Euler's method to perform the \texttt{UVA-step} implemented with Jax in order to propagate derivatives.
To optimize the parameters $\btheta$ and $\blambda$ we used \texttt{adam} \citep{kingma2014adam} with a step size of 1e-4.  
A single Monte Carlo sample was found to adequately estimate the stochastic gradient for each update, which incorporates all of the training observations.
We stop iterating the algorithm when the loss stops improving for 500 iterations --- this typically occurred after 10-15{,}000 iterations.